%% file: main.tex
\newif\ifsubmission \newif\ifpaperonly \newif\ifarxiv \arxivtrue\RequirePackage{fix-cm}
\documentclass[10pt,twocolumn,letterpaper]{article}
\usepackage[pagenumbers]{cvpr}\makeatletter\providecommand{\LenToUnit}{}\renewcommand{\LenToUnit}[1]{\strip@pt\dimexpr#1*\p@/\unitlength}\makeatother
\usepackage{times,graphicx,amsmath,amssymb}
\usepackage{mathptmx}
\newcommand{\given}{\mid\nobreak}

\usepackage{caption}
\makeatletter\@twocolumnfalse\makeatother
\usepackage{multicol}
\makeatletter\@twocolumntrue\makeatother
\usepackage[pagebackref,breaklinks,hidelinks]{hyperref}
\makeatletter
\patchcmd{\NAT@sort@cites}{\expandafter\immediate\expandafter\write}{\expandafter\write}{}%
  {\PackageError{main}{natbib patch failed: citations would be numbered in source order}{}}
\patchcmd{\NAT@citexnum}{\def@NAT@last@yr{--\NAT@penalty}}{\def@NAT@last@yr{\mbox{--}\nobreak}}{}%
  {\PackageError{main}{natbib patch failed: citation ranges could break at their dash}{}}
\patchcmd{\NAT@citexnum}{\NAT@parse{\@citeb}}{\NAT@parse{\@citeb}\Hy@backout{\@citeb}}{}%
  {\PackageError{main}{natbib patch failed: keys inside citation ranges would get no back-references}{}}
\makeatother
\newif\ifcitemargin \citemargintrue \citemarginfalse\makeatletter\providecommand\CM@entry[4]{}\makeatother
\def\confName{CVPR}\def\confYear{2027}
\hypersetup{pdftitle={Reconstructing Natural Images from Brain Activity during Natural Vision},pdfauthor={Tejasvi Kothapalli; Ryan Ressmeyer; Declan Rowley; Gregory D. Horwitz; Jude Mitchell; Alexander C. Huk; Bruno Olshausen; Jacob Yates},pdfsubject={Authored figure and abstract working draft}}
\hypersetup{pdfsubject={}}\title{Reconstructing Natural Images from Brain Activity during Natural Vision}
\newcommand{\AuthorBlock}{\begin{minipage}[t]{\textwidth}\centering\small
Tejasvi Kothapalli$^{1,*}$\quad Ryan Ressmeyer$^{2,*}$\quad Declan Rowley$^{3,*}$\\
Gregory D. Horwitz$^{2}$\quad Jude Mitchell$^{4}$\quad Alexander C. Huk$^{3}$\quad Bruno Olshausen$^{1}$\quad Jacob Yates$^{1}$\par
\vspace{3pt}\footnotesize
$^1$UC Berkeley\quad $^2$University of Washington\quad $^3$UCLA\quad $^4$University of Rochester\quad $^*$Equal contribution
\end{minipage}}
\author{\AuthorBlock}\makeatletter
\patchcmd{\@maketitle}{\vskip .375in}{\vspace*{-7pt}}{}{\PackageError{main}{Title patch failed}{}}
\patchcmd{\@maketitle}{\vspace*{24pt}}{\vspace*{8pt}}{}{\PackageError{main}{Title patch failed}{}}
\patchcmd{\@maketitle}{\vspace*{12pt}}{\vspace*{2pt}}{}{\PackageError{main}{Title patch failed}{}}
\makeatother
\input{text/captions}

\newcommand{\FigRefAt}[3]{\put(#1,#2){\makebox(0,0)[t]{\fontfamily{phv}\fontsize{5.6}{5.6}\selectfont\color[gray]{.42}\cite{#3}}}}
\newlist{outline}{itemize}{2}
\setlist[outline,1]{label=\textbullet,leftmargin=0.9em,labelsep=0.4em,nosep,topsep=1pt}
\setlist[outline,2]{label=--,leftmargin=0.9em,labelsep=0.4em,nosep}

\makeatletter

\renewcommand\paragraph{\@startsection{paragraph}{4}{\z@}{0pt}{-1em}{\normalfont\normalsize\bfseries}}
\begin{document}
\onecolumn
\@maketitle
\noindent\includegraphics[width=\linewidth]{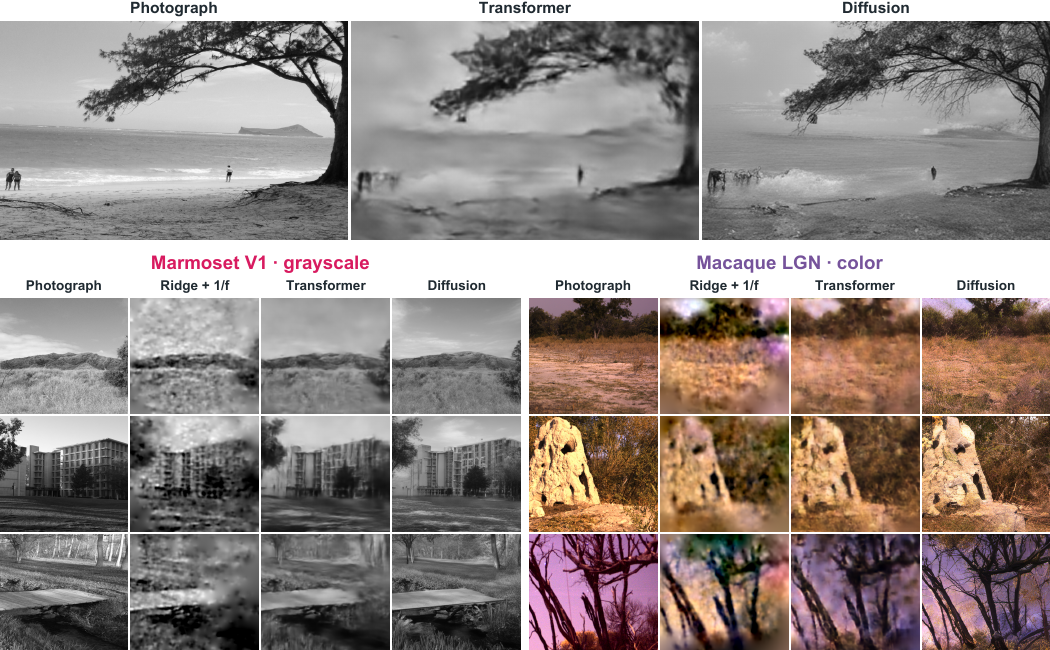}\par{\captionof{figure}{\CaptionOne}}
\vspace{10pt}{\setlength{\multicolsep}{0pt}\begin{multicols}{2}
\centerline{\large\bf Abstract}\vspace*{12pt}%
{\noindent\it\ignorespaces\input{text/abstract}\looseness=-1\par}
\end{multicols}}
\clearpage
\makeatother
\twocolumn[{\noindent\begin{minipage}{\textwidth}\noindent\includegraphics[width=\linewidth]{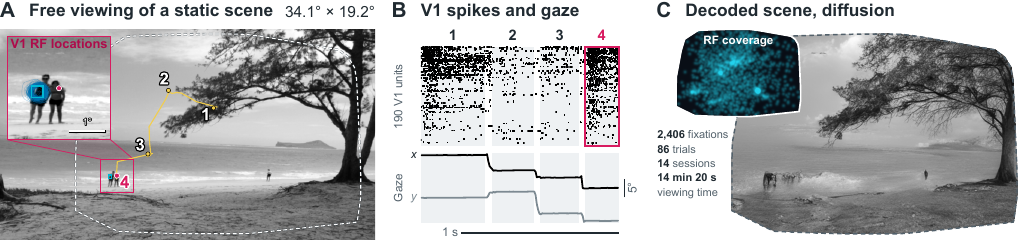}\par{\captionof{figure}{\CaptionTwo}}\end{minipage}\par\vspace{\textfloatsep}}]
\section{Introduction}\label{sec:intro}
\input{text/introduction}
\section{Background and related work}\label{sec:related}
\input{text/related_work}
\begin{figure*}[t]\noindent\includegraphics[width=\linewidth]{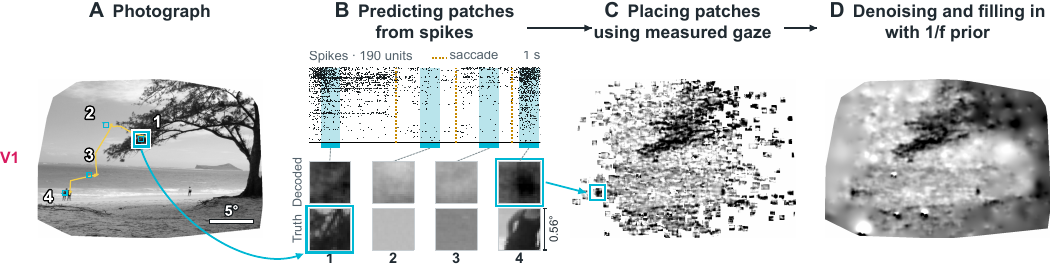}\par{\captionof{figure}{\CaptionThree}}\end{figure*}
\setcounter{dbltopnumber}{1}
\setlength{\dbltextfloatsep}{12pt plus 2pt minus 2pt}
\section{Methods}\label{sec:methods}
\begin{figure*}[t]\noindent\includegraphics[width=\linewidth]{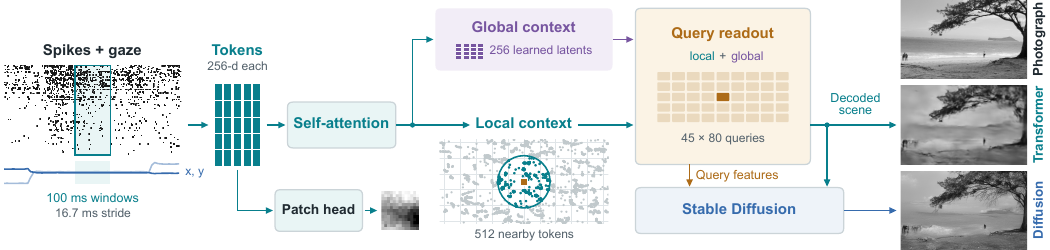}\par{\captionof{figure}{\CaptionFour}}\end{figure*}
\begin{figure*}[t]\noindent\includegraphics[width=\linewidth]{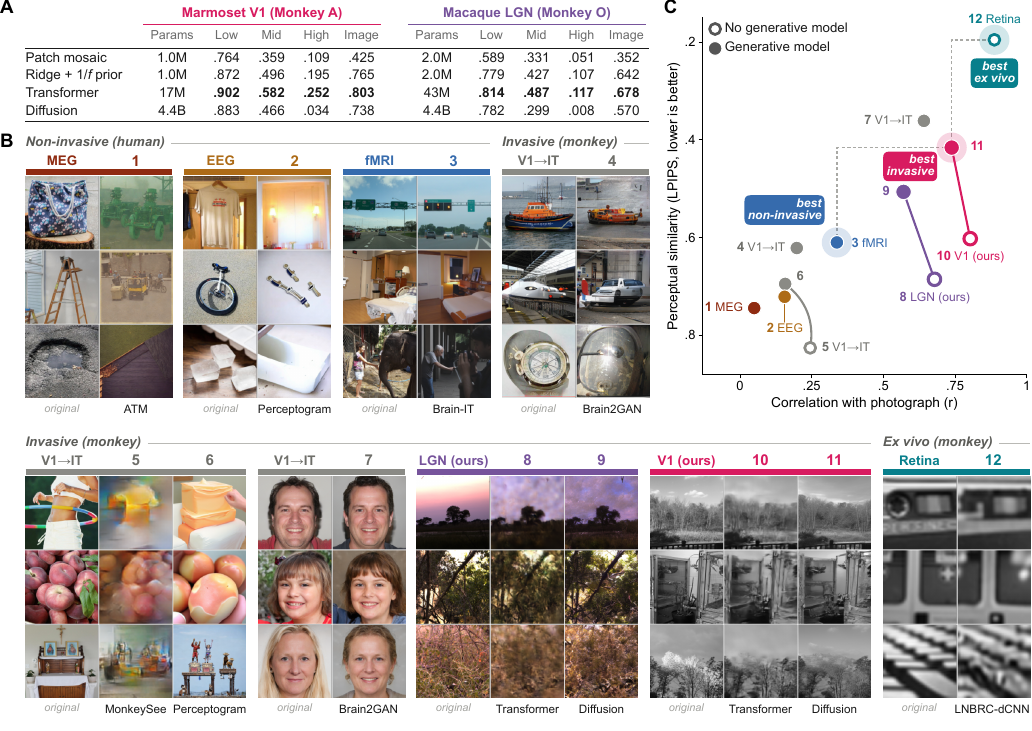}\llap{\makebox[\linewidth][l]{\setlength{\unitlength}{\linewidth}\begin{picture}(0,0)\input{text/fig5_refs}\end{picture}}}\par{\captionof{figure}{\CaptionMethods}}\end{figure*}
\input{text/methods}
\newsavebox{\GridBox}
\begin{figure*}[t]\noindent\includegraphics[width=\linewidth]{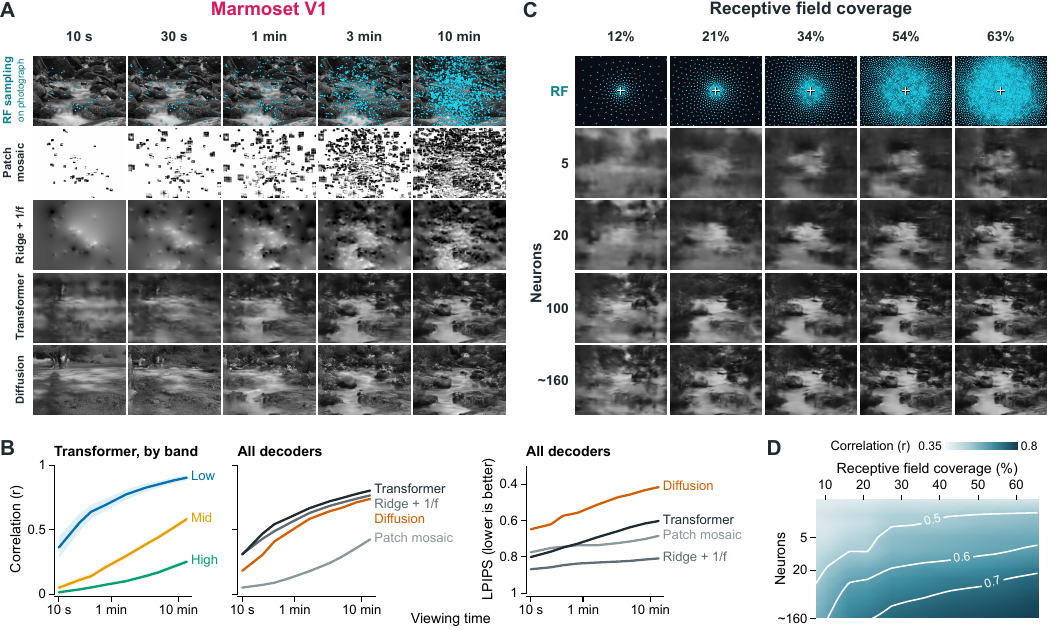}\par{\captionof{figure}{\CaptionFive}}\vspace{-1pt}\end{figure*}
\begin{figure*}[t]\noindent\includegraphics[width=\linewidth]{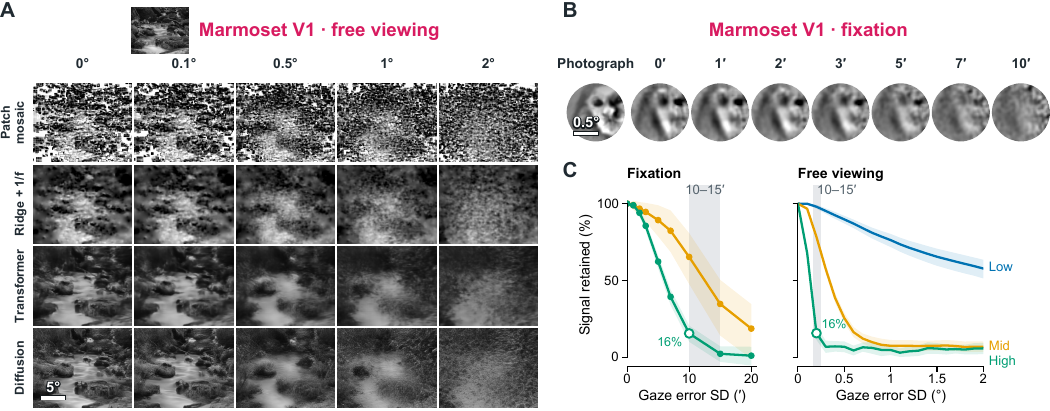}\par{\captionof{figure}{\CaptionSeven}}\vspace{-4pt}\end{figure*}
\input{text/results}
\input{text/discussion}
\onecolumn
\raggedbottom
\begingroup\begin{multicols}{2}
\iftoggle{cvprfinal}{%
\section*{Acknowledgements}
TK thanks the faculty and TAs at the Cold Spring Harbor Computational Neuroscience:\ Vision 2026 Summer Course and the MIT Brains, Minds and Machines 2026 Summer Course. Conversations with Eero Simoncelli, E.J.~Chichilnisky, Jim DiCarlo, Gabriel Kreiman, Jack \mbox{Gallant}, and Michael Silver were particularly helpful. TK would also like to thank Ashwin Nathan for access to \mbox{OpenAI's} Codex and Andi Gu for helpful conversations about training large models.
\paragraph{Funding.} This research was supported by the NIH (R00 EY032179 to JLY; R01 EY030998 to JFM; R01 EY037680 to JLY and GDH), NEI K99-EY037839-01 to DPR, an NSF Graduate Research Fellowship (DGE-2140004) to RAR, the Fuster Endowment to ACH and a CIVO fellowship to TK. TK also receives NIH/NEI training support (T32 EY007043).
}{}
\section*{LLM Usage Statement}
OpenAI's Codex and Anthropic's Claude Code were used at every step of this project: ideation, data preprocessing, building the architecture, coding, hyperparameter testing, running the models, creating the figures, and writing the paper. The authors were closely involved at every step, from collecting the data from the primates to data preprocessing, creating the method, and writing the paper.
{\small
\setlength{\bibsep}{0.5pt plus 1pt minus 0.5pt}
\bibliographystyle{ieeenat_fullname_citeorder}
\bibliography{references}}
\end{multicols}\endgroup
\let\next\relax\next
\clearpage
\setcounter{figure}{0}\renewcommand{\thefigure}{S\arabic{figure}}\renewcommand{\theHfigure}{S\arabic{figure}}
\setcounter{table}{0}\renewcommand{\thetable}{S\arabic{table}}\renewcommand{\theHtable}{S\arabic{table}}
\section*{Supplementary Material}\label{supp:start}
\newcommand{\TitlePriorTail}{Image Prior}\newcommand{\TitleFill}{Filling in Missing Pixels}\newcommand{\TitlePatches}{Reconstructing a Scene from Decoded Patches}
\raggedbottom
\appendix
\input{text/supp_nogaze}
\clearpage
\section{Macaque LGN Figures}
\label{app:lgn-figures}
Figures of the paper that show only marmoset V1 are repeated here for the macaque LGN.
\par\vspace{6pt}
\noindent\includegraphics[width=\linewidth]{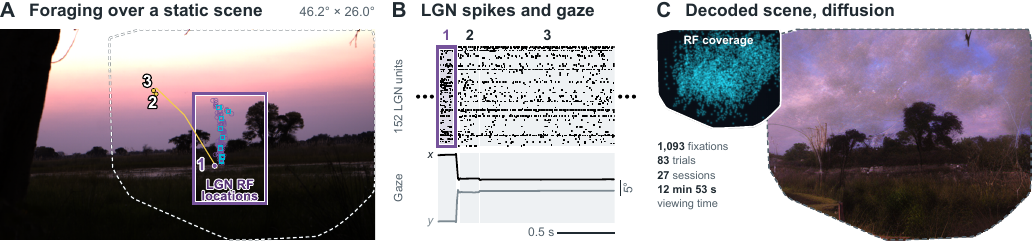}\par{\captionof{figure}{\CaptionSThree\label{figS:lgn-decoding}}}
\vspace{10pt}
\noindent\includegraphics[width=\linewidth]{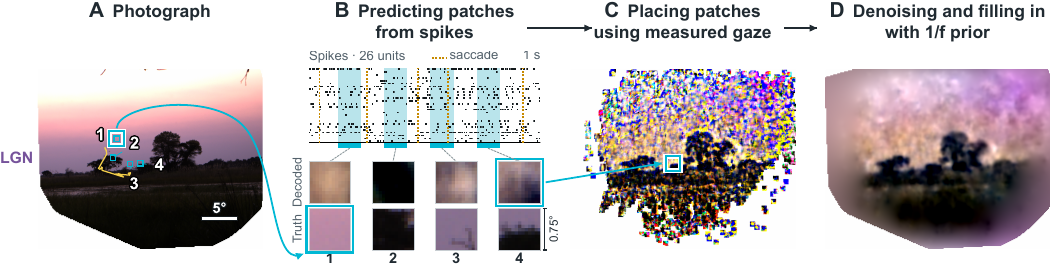}\par{\captionof{figure}{\CaptionSFour\label{figS:lgn-ridge}}}
\clearpage
\noindent\includegraphics[width=\linewidth]{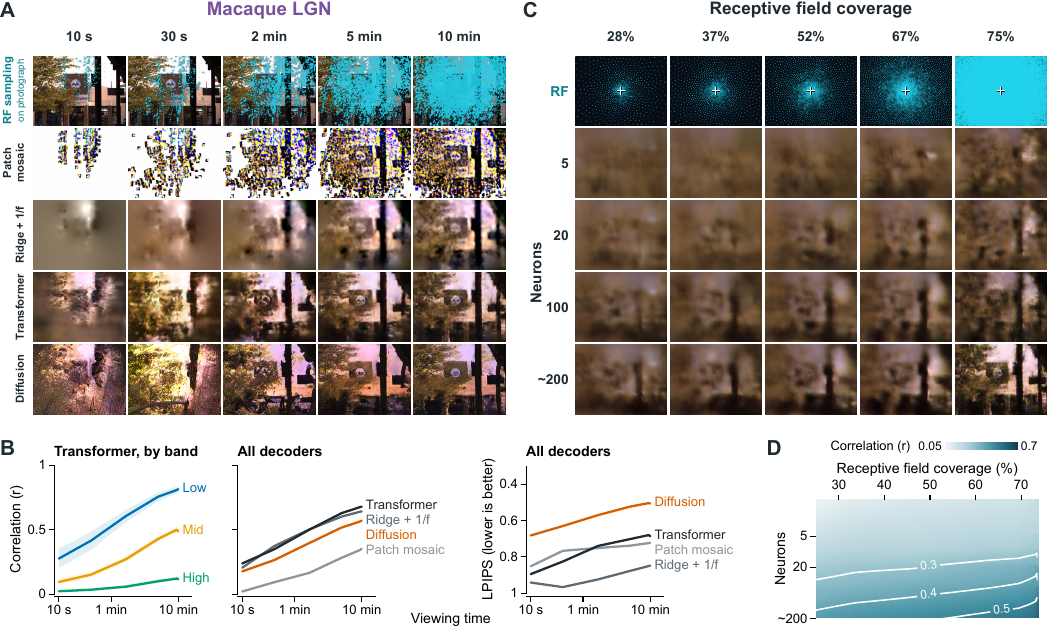}\par{\captionof{figure}{\CaptionSLGNViewing\label{figS:lgn-viewing}}}
\clearpage
\noindent\includegraphics[width=\linewidth]{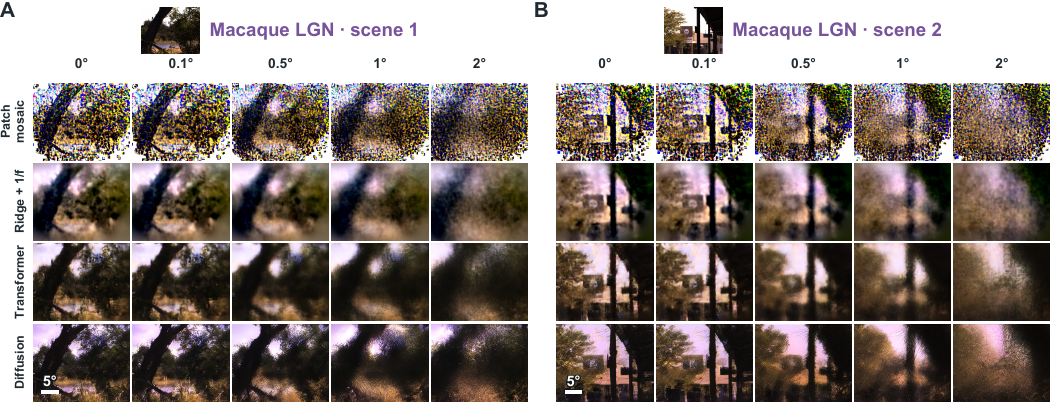}\par\vspace{8pt}
\noindent\begin{minipage}[t]{0.5\linewidth}\vspace{0pt}\includegraphics[width=\linewidth]{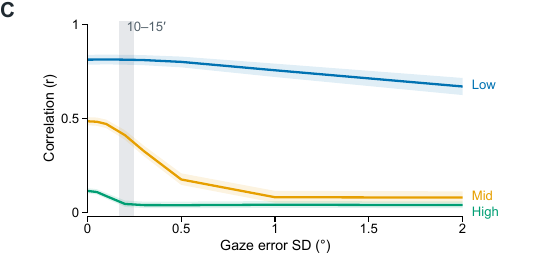}\end{minipage}\hfill
\begin{minipage}[t]{0.47\linewidth}\vspace{0pt}\captionof{figure}{\CaptionSLGNGaze\label{figS:lgn-gaze}}\end{minipage}
\clearpage
\input{text/appendix_comparison}
\clearpage
\noindent\includegraphics[width=\linewidth]{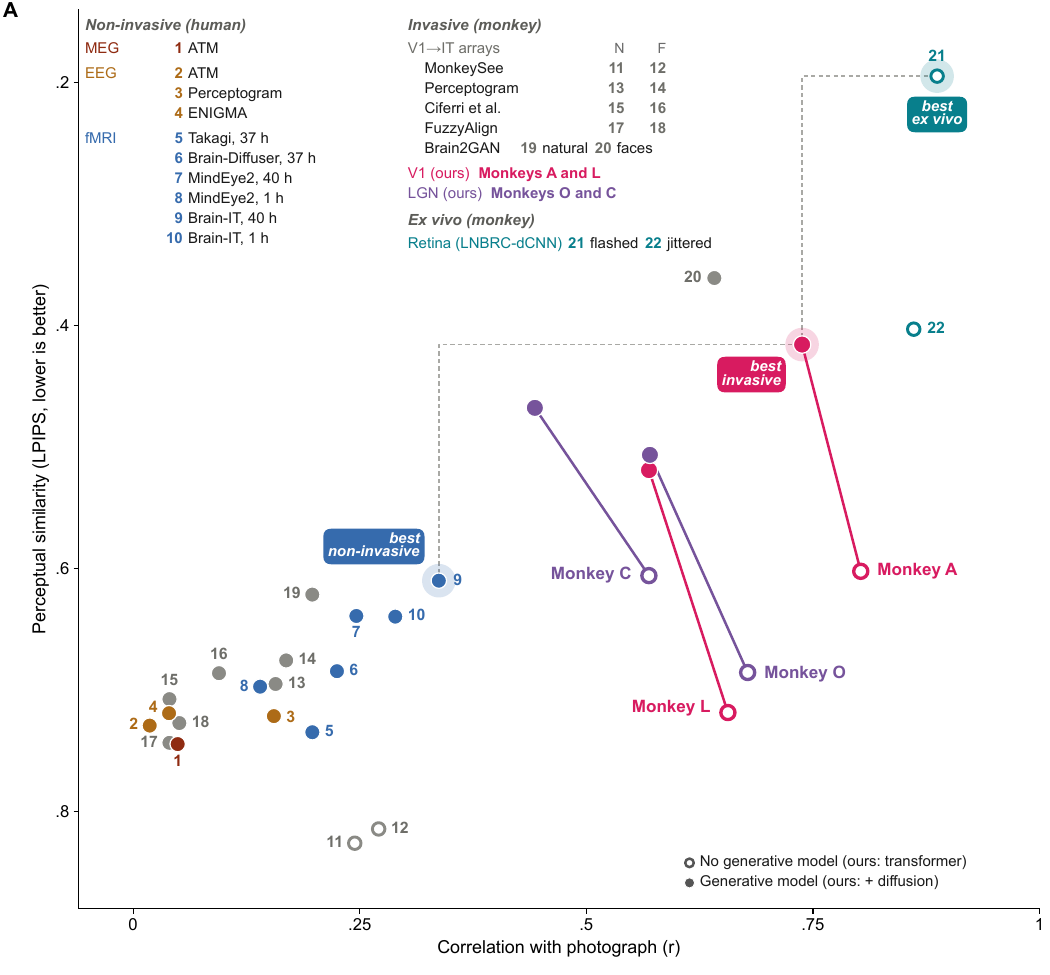}\par{\captionof{figure}{\CaptionSLandscape\label{figS:landscape}}}
\clearpage
\noindent\includegraphics[width=\linewidth]{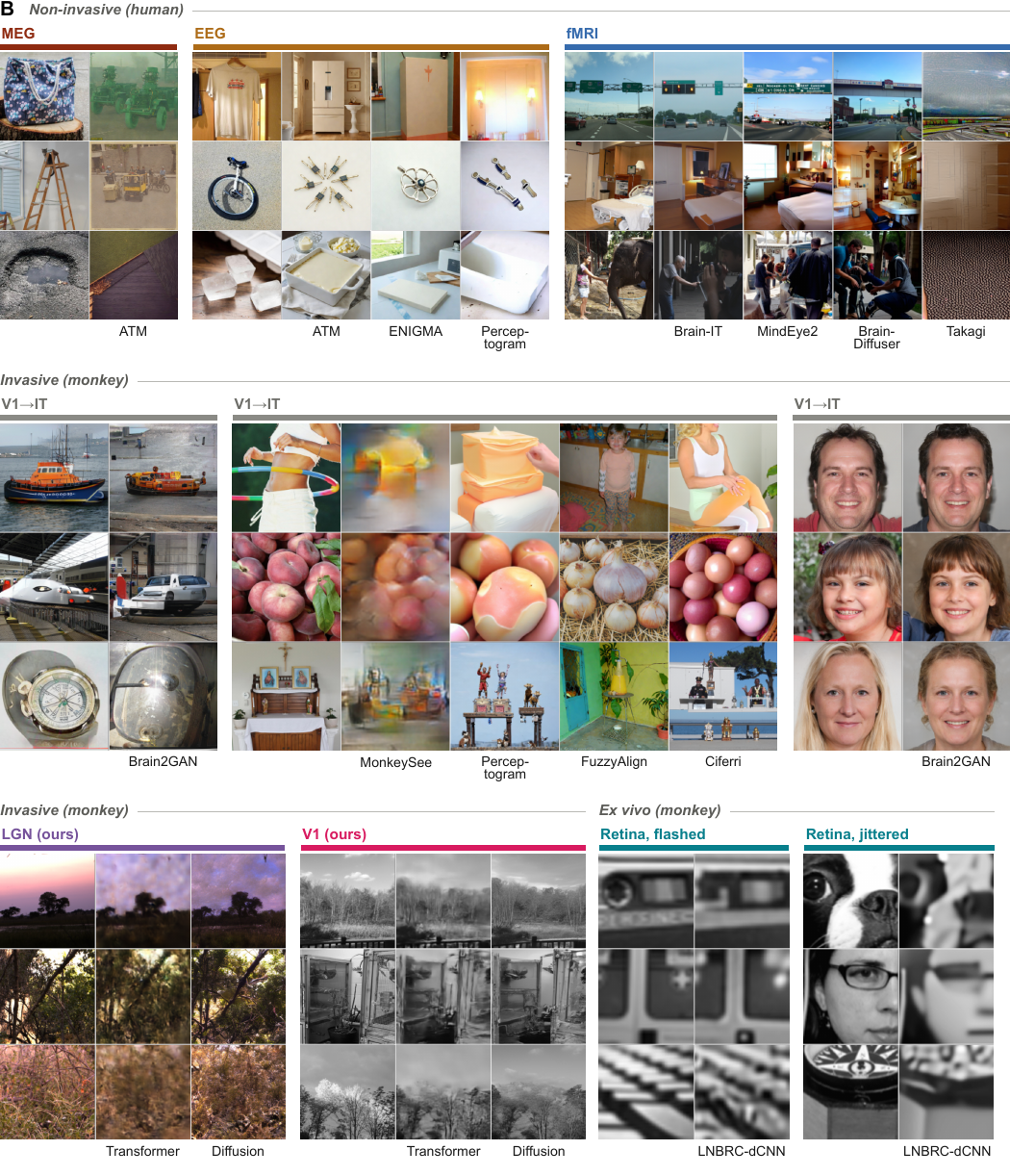}\par\vspace{5pt}{\small\noindent\textbf{Figure~\thefigure.}~(continued)\par}
\clearpage
{\centering\captionof{table}{\CaptionTMetrics\label{tabS:metrics}}\vspace{4pt}
\noindent\includegraphics[width=\linewidth]{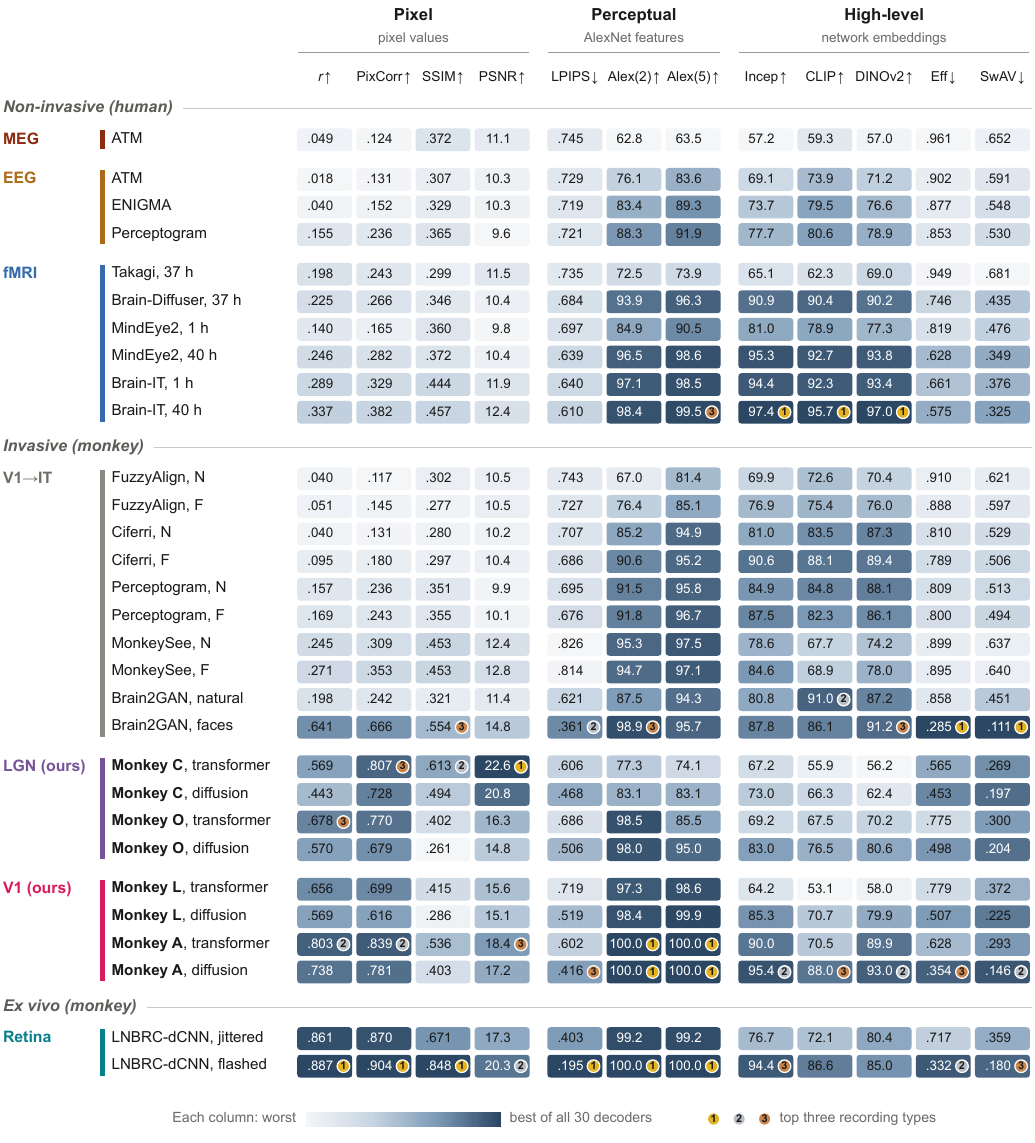}\par}
\clearpage
{\centering\captionof{table}{\CaptionTData\label{tabS:data}}\vspace{4pt}
\noindent\includegraphics[width=\linewidth]{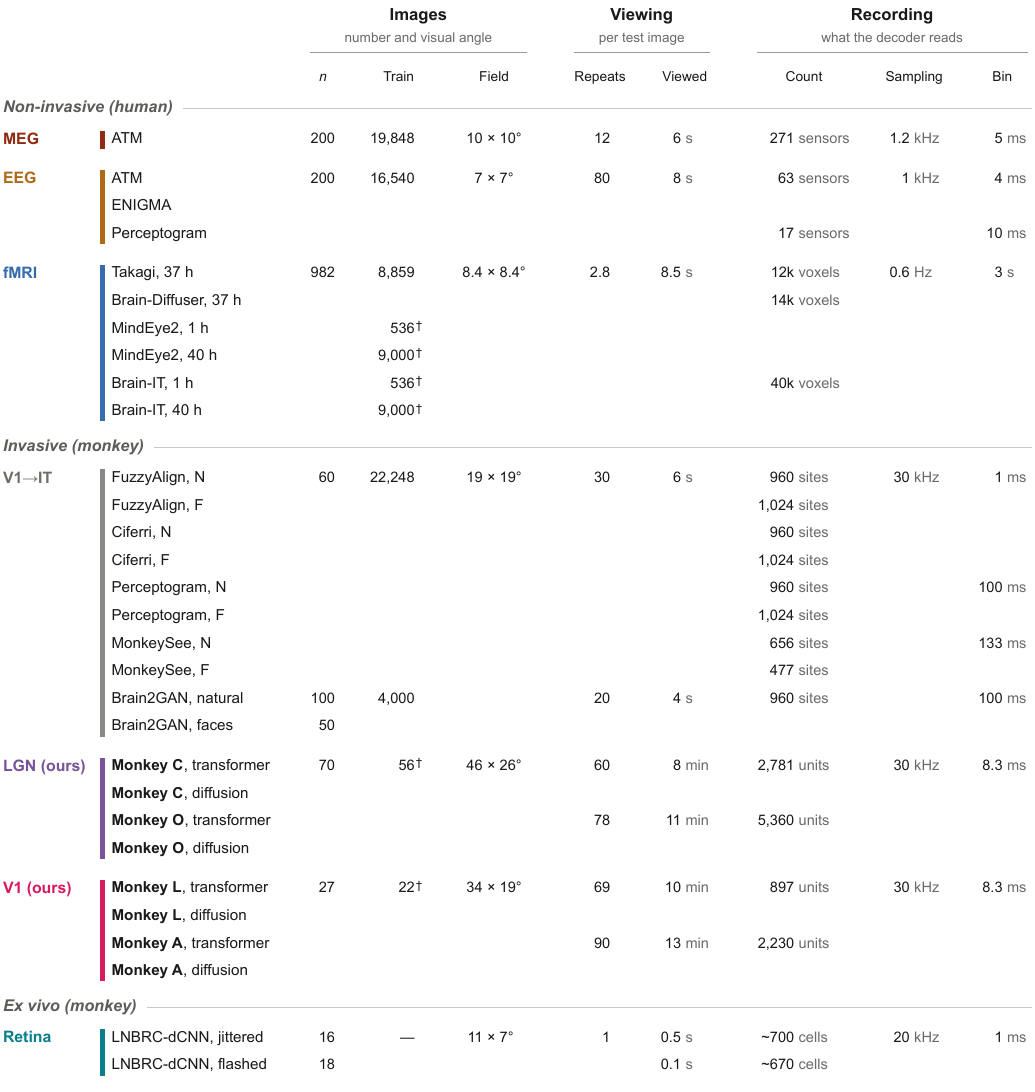}\par}
\clearpage
\noindent\includegraphics[width=\linewidth]{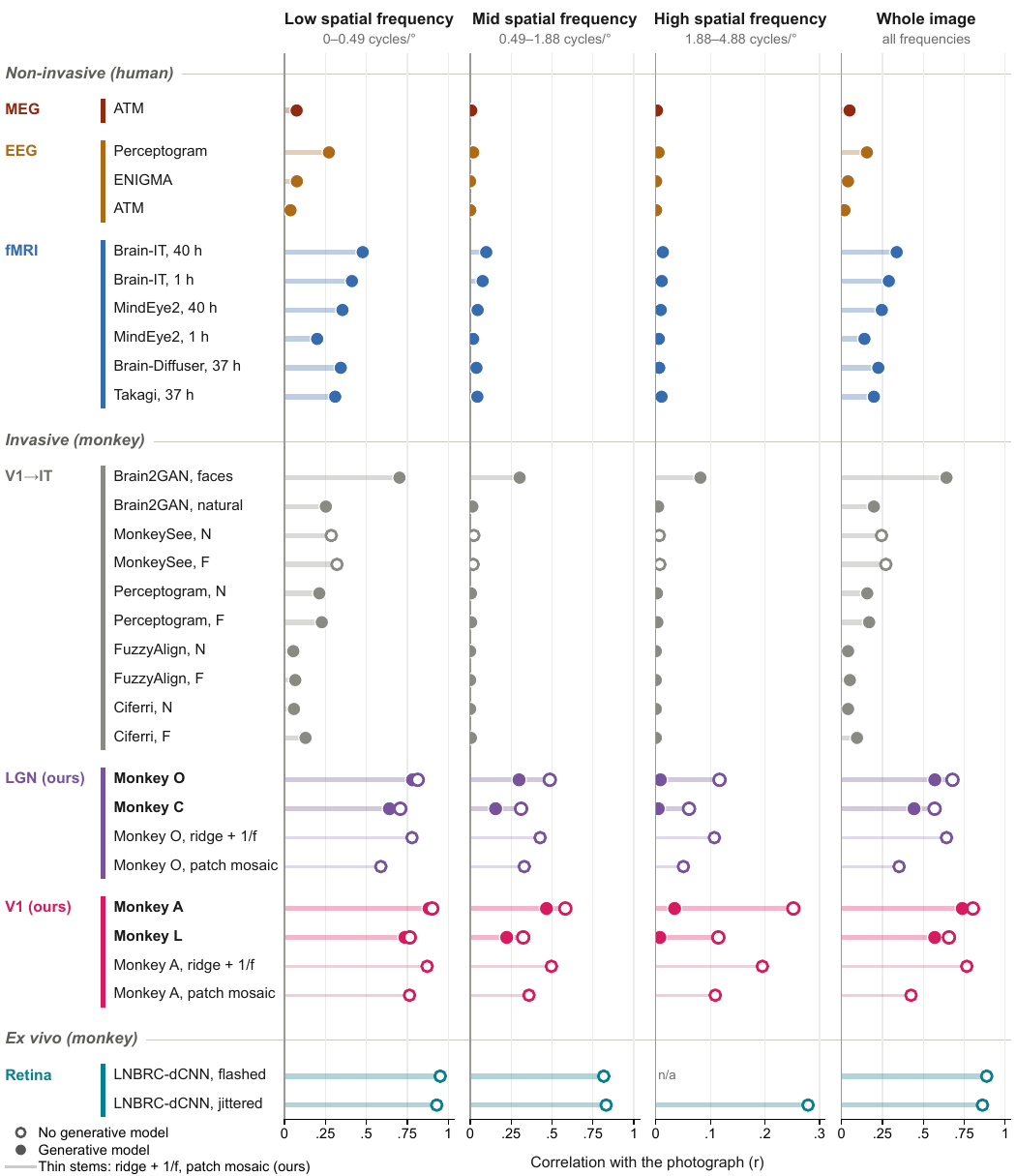}\par{\captionof{figure}{\CaptionSBands\label{figS:bands}}}
\clearpage
\section{More Fixation Examples}
\label{app:fixation-examples}
\noindent\includegraphics[width=\linewidth]{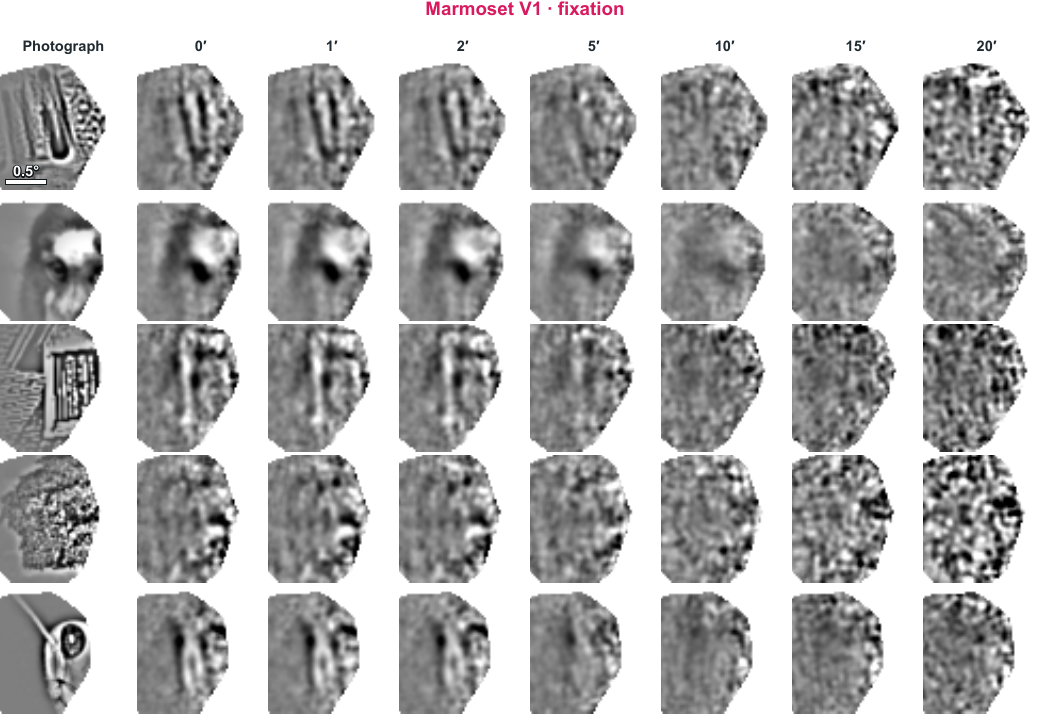}\par{\captionof{figure}{\CaptionSFixation\label{figS:fixation}}}
\clearpage
\input{text/appendix_prior}
\clearpage
\input{text/appendix_recordings}
\input{text/appendix_offset}
\clearpage
\section{Every Held-Out Scene}
\label{app:scenes}
\noindent\includegraphics[width=\linewidth]{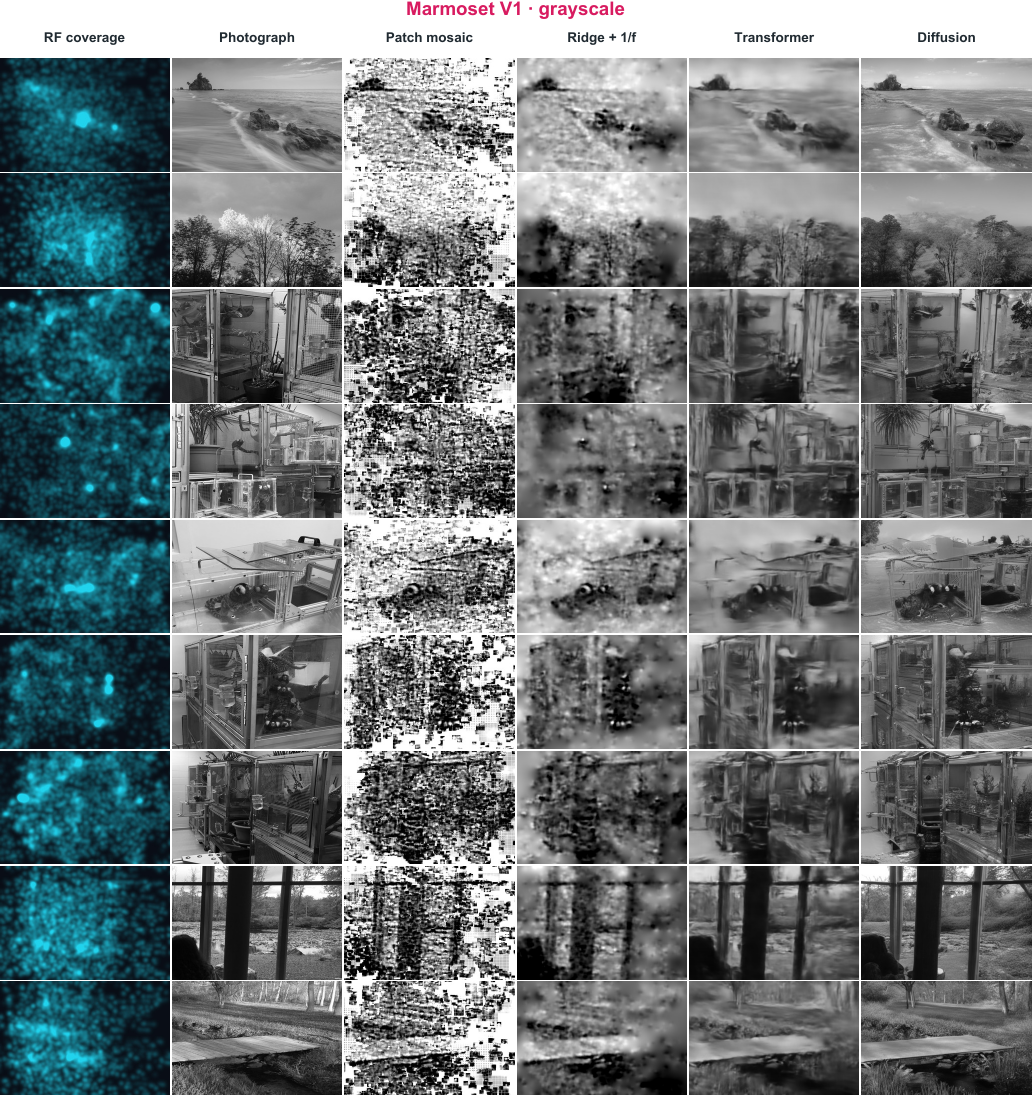}\par{\captionof{figure}{\CaptionSOne\label{figS:v1-scenes}}}
\clearpage
\noindent\includegraphics[width=\linewidth]{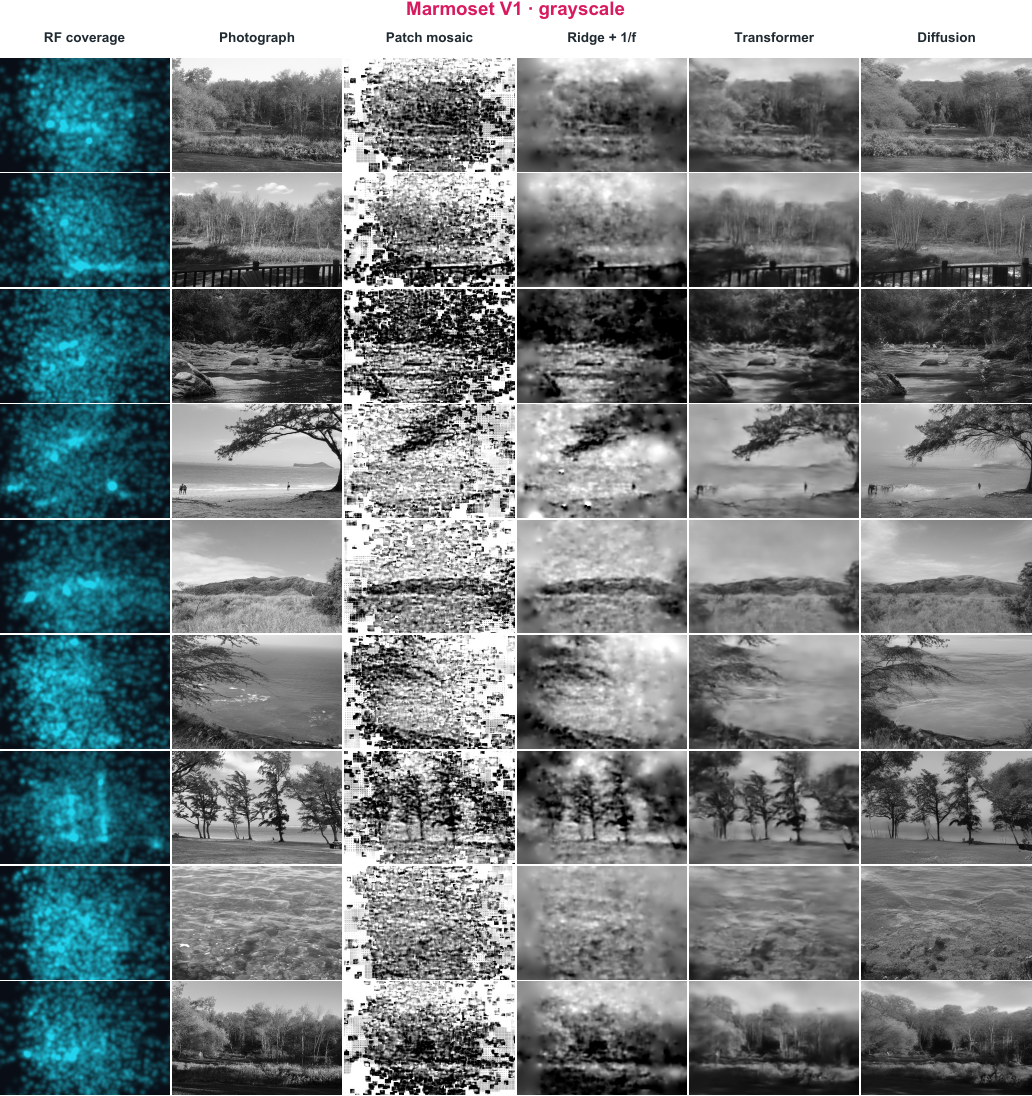}\par\vspace{5pt}{\small\noindent\textbf{Figure~\thefigure.}~(continued)\par}
\clearpage
\noindent\includegraphics[width=\linewidth]{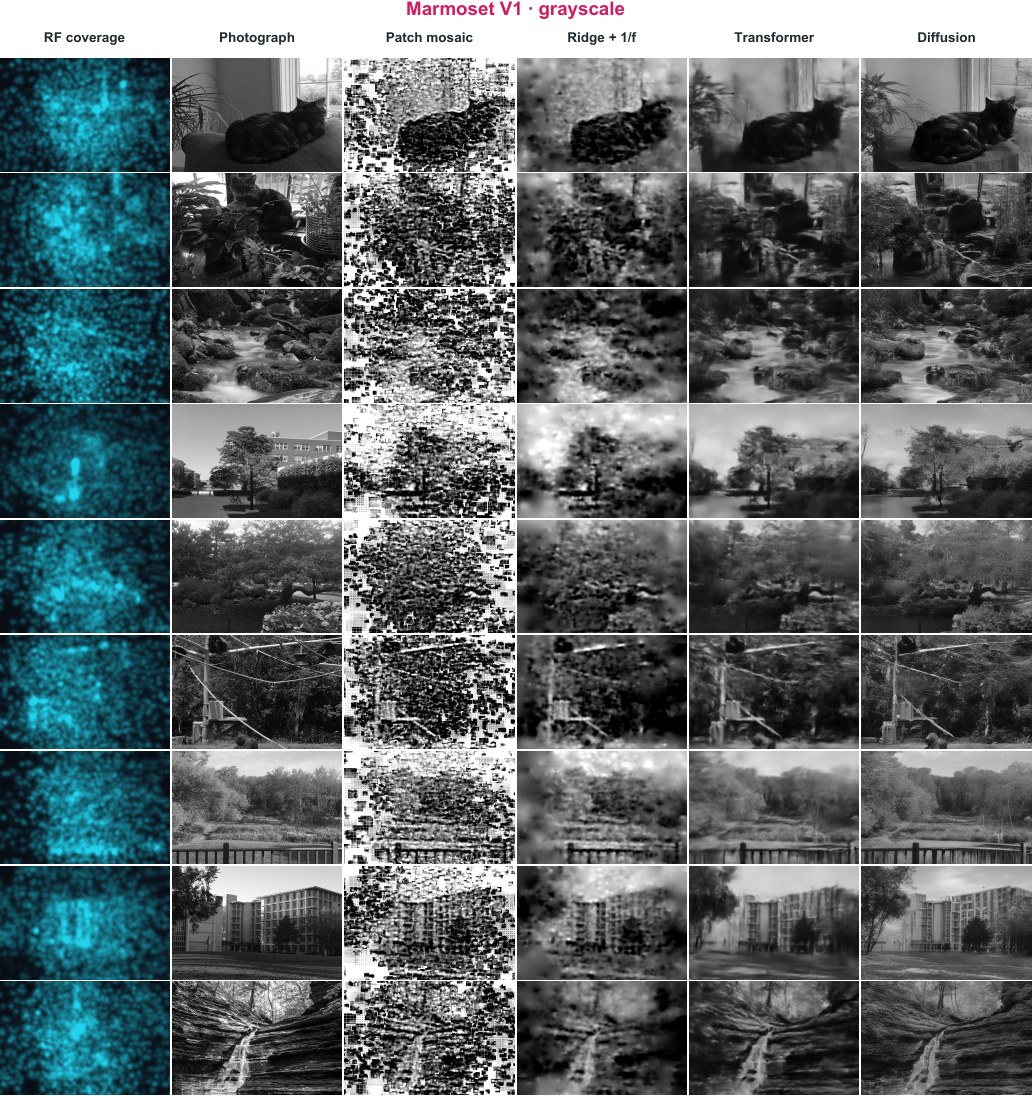}\par\vspace{5pt}{\small\noindent\textbf{Figure~\thefigure.}~(continued)\par}
\clearpage
\noindent\includegraphics[width=\linewidth]{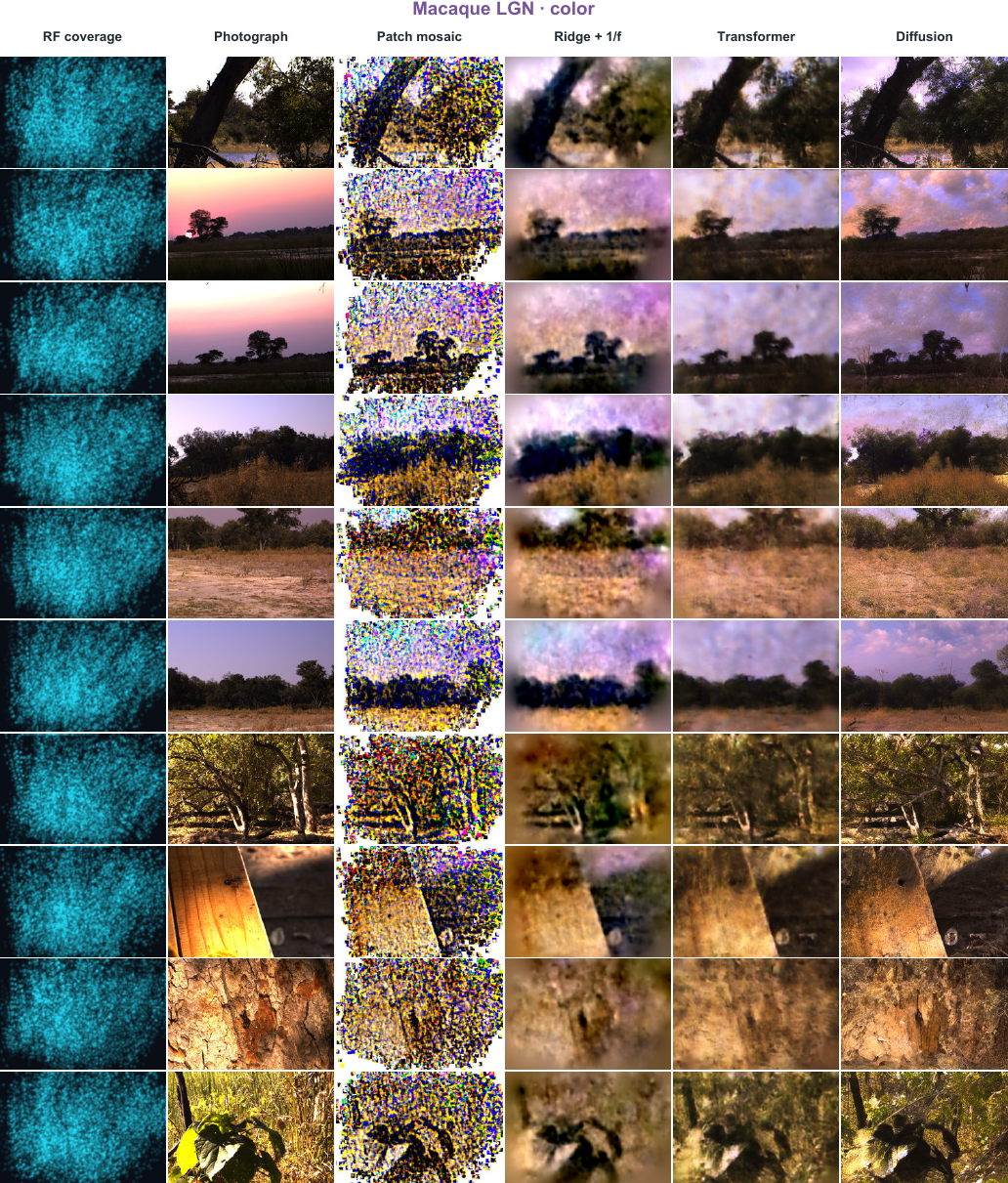}\par{\captionof{figure}{\CaptionSTwo\label{figS:lgn-scenes}}}
\clearpage
\noindent\includegraphics[width=\linewidth]{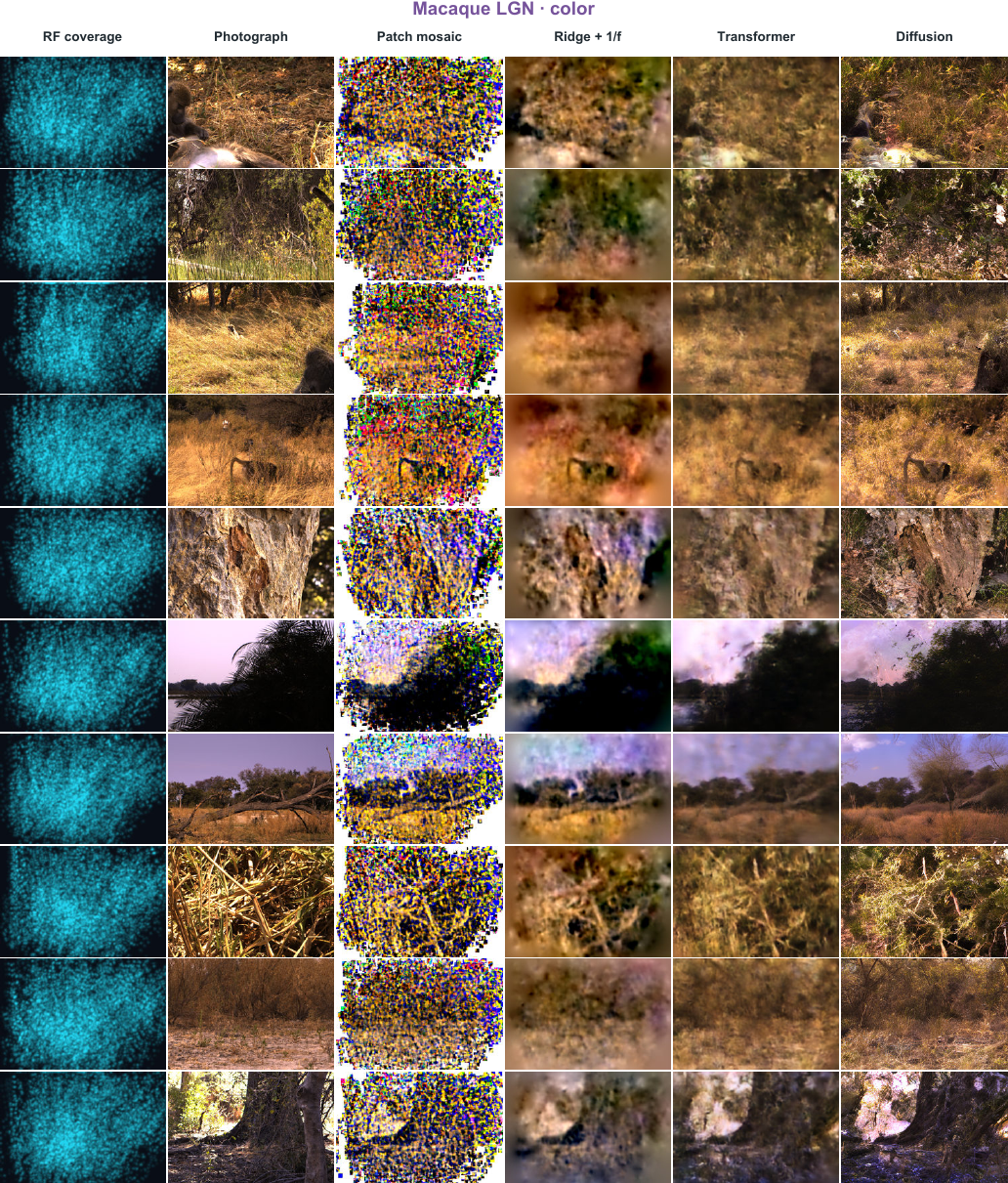}\par\vspace{5pt}{\small\noindent\textbf{Figure~\thefigure.}~(continued)\par}
\clearpage
\noindent\includegraphics[width=\linewidth]{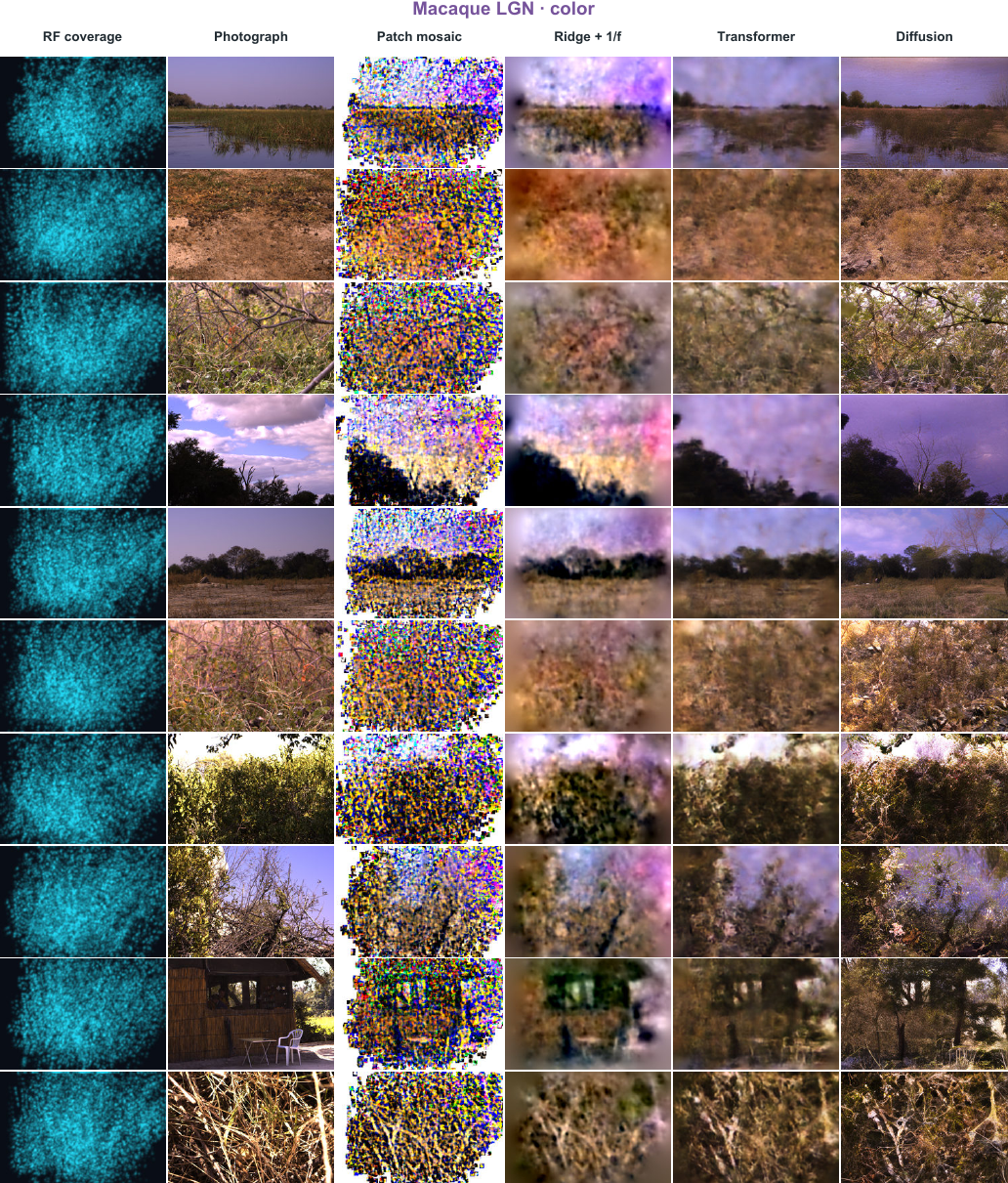}\par\vspace{5pt}{\small\noindent\textbf{Figure~\thefigure.}~(continued)\par}
\clearpage
\noindent\includegraphics[width=\linewidth]{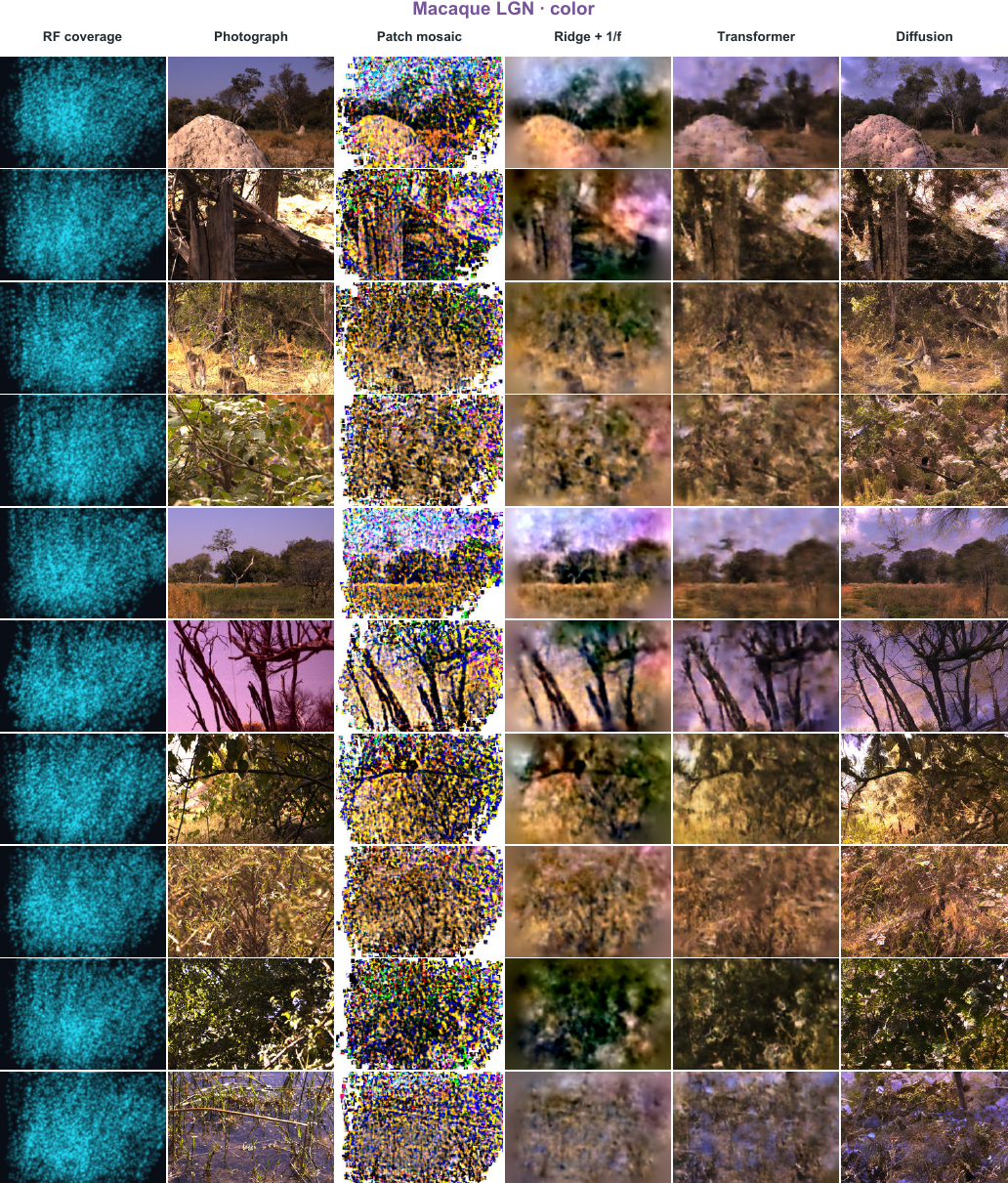}\par\vspace{5pt}{\small\noindent\textbf{Figure~\thefigure.}~(continued)\par}
\clearpage
\noindent\includegraphics[width=\linewidth]{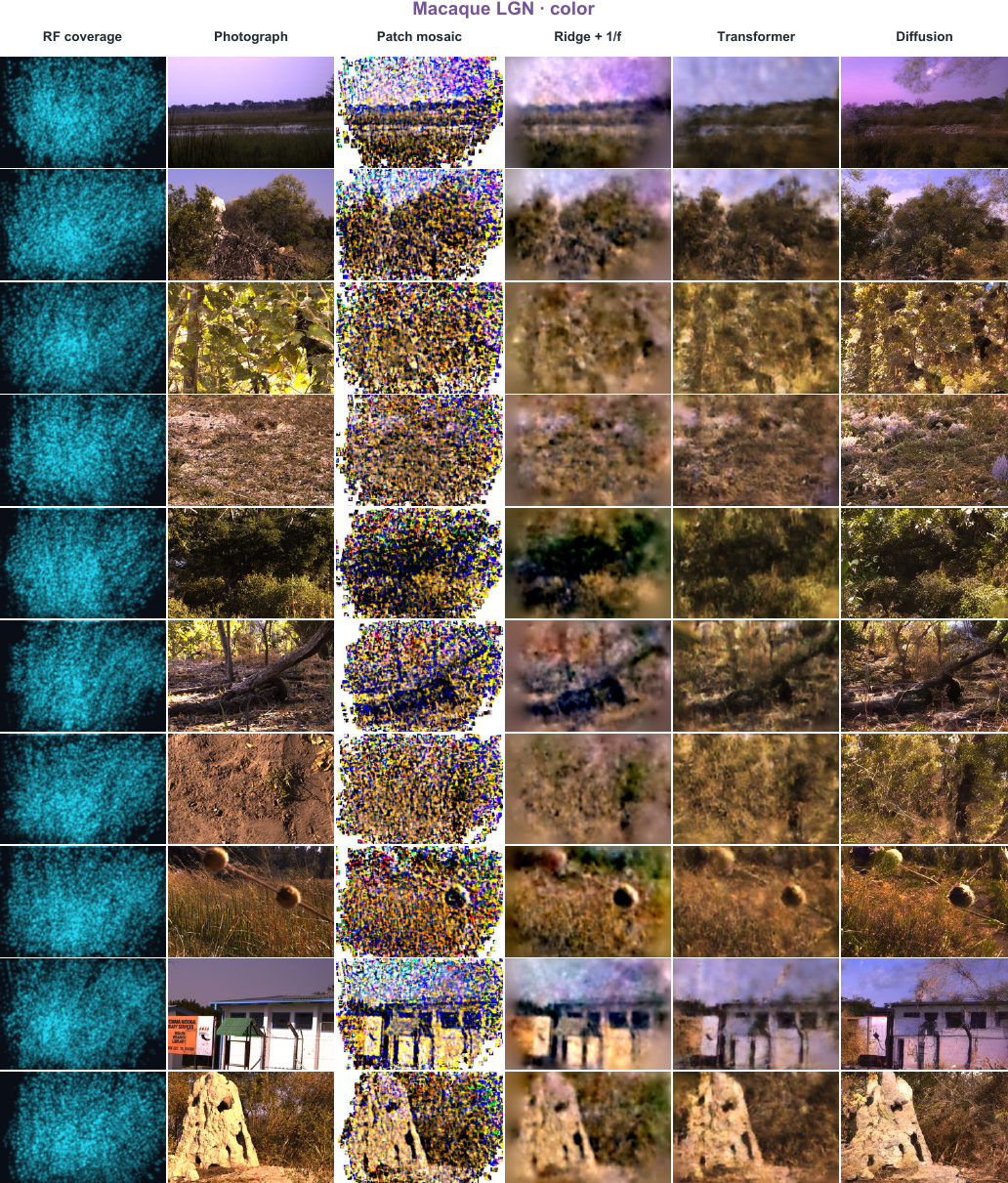}\par\vspace{5pt}{\small\noindent\textbf{Figure~\thefigure.}~(continued)\par}
\clearpage
\noindent\includegraphics[width=\linewidth]{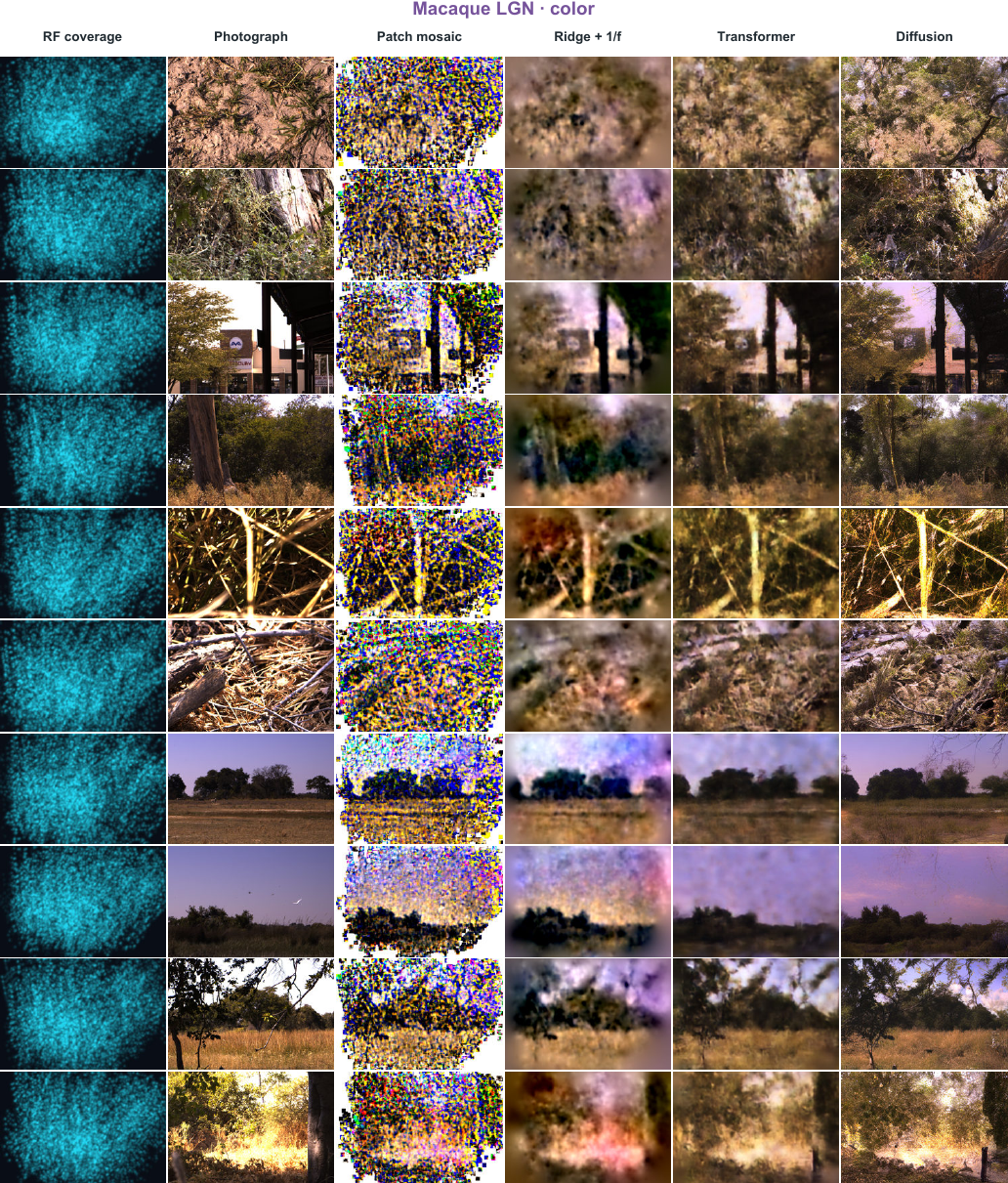}\par\vspace{5pt}{\small\noindent\textbf{Figure~\thefigure.}~(continued)\par}
\clearpage
\noindent\includegraphics[width=\linewidth]{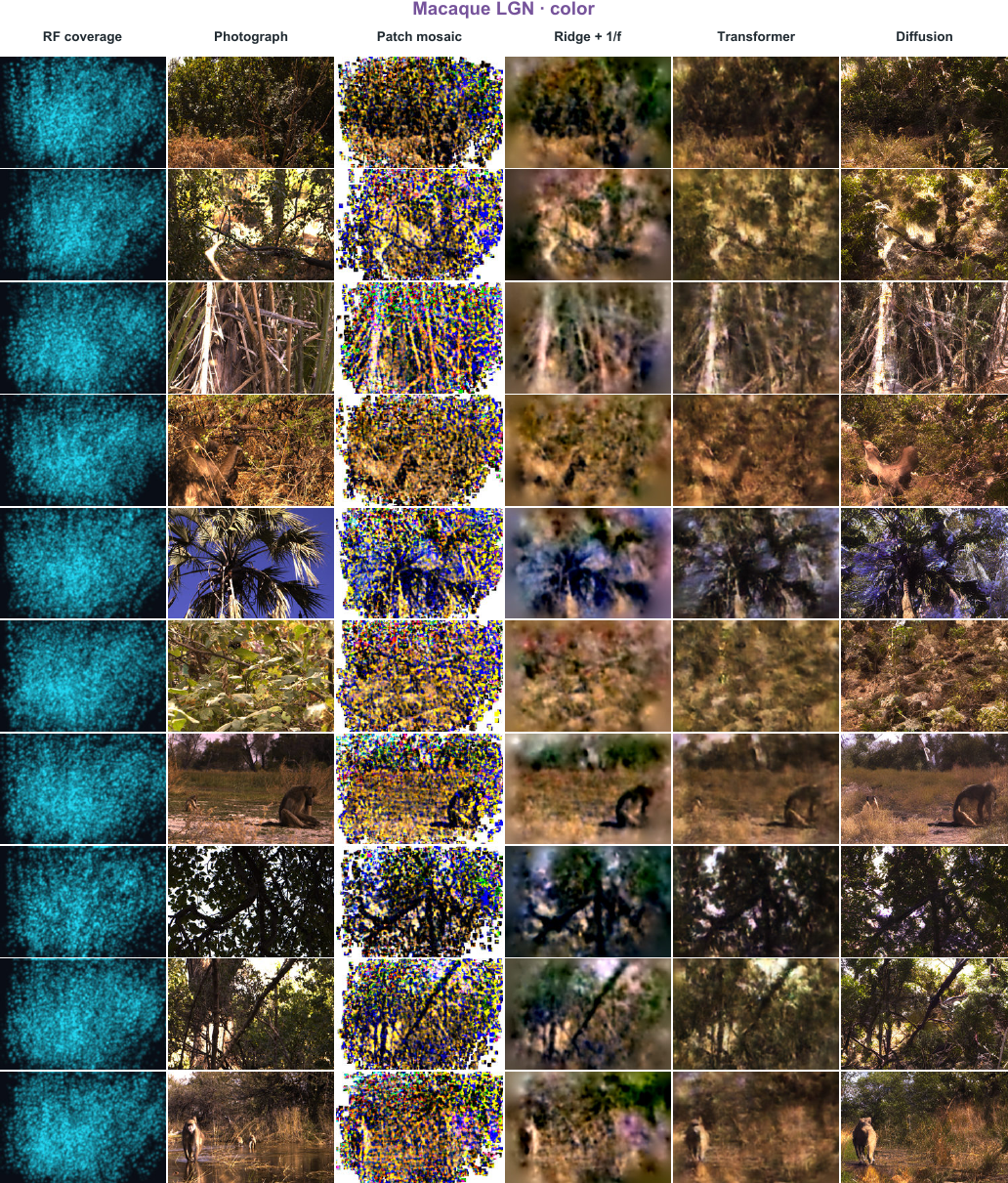}\par\vspace{5pt}{\small\noindent\textbf{Figure~\thefigure.}~(continued)\par}
\end{document}

%% file: text/captions.tex
\newcommand{\CaptionOne}{\textbf{Image reconstructions from primate neural activity during free-viewing.} These images are reconstructed by combining neural activity across eye movements. The top row shows one marmoset V1 scene, enlarged. Columns show the photograph and its reconstructions by ridge regression with a 1/$f$ prior, the transformer, and diffusion. Figs.~\ref{figS:v1-scenes} and~\ref{figS:lgn-scenes} show every held-out scene.}
\newcommand{\CaptionTwo}{\textbf{Decoding during free viewing.} \textbf{A},~An example eye trace (yellow) of a marmoset viewing a static scene, with V1 receptive fields (RFs). The inset zooms in on fixation~4 to show their scale. The light blue circles are the overlaid RFs of 190 V1 units. The cyan square is the patch the models predict. It sits at a fixed offset from gaze (red dot). \textbf{B},~Spikes of 190 V1 units (rows) and gaze over time. Each jump in gaze is a saccade. Fixations are numbered as in A. The box marks fixation~4 (inset in A). \textbf{C},~The scene decoded from V1 by the diffusion model. The inset shows where V1 RFs landed over all fixations, trials and sessions. Fig.~\ref{figS:lgn-decoding} shows the same for macaque LGN.}
\newcommand{\CaptionSOne}{\textbf{Every held-out scene, marmoset V1.} All 27 photographs, in six columns: where the receptive fields sampled the scene over all viewing time (brighter where they sampled more), photograph, patch mosaic (no prior), ridge regression with a 1/$f$ prior, transformer, and diffusion. Each row shows the largest rectangle inside the part of the scene that the receptive fields covered. Fig.~1 shows four of them, and Fig.~\ref{figS:bands} averages over all 27. Scenes are in alphabetical order of file name, not sorted by quality.}%
\newcommand{\CaptionSTwo}{\textbf{Every held-out scene, macaque LGN.} All 70 photographs, in the columns of Fig.~\ref{figS:v1-scenes} and in the order of the stimulus set. Fig.~1 shows three of them, and Fig.~\ref{figS:bands} averages over all 70.}
\newcommand{\CaptionSThree}{\textbf{Decoding from macaque LGN.} \textbf{A},~An example eye trace (yellow) of a macaque searching a static scene for small targets. The purple ellipses are the receptive fields (RFs) of the 152 LGN units in B, drawn at fixation~1. Unlike V1 RFs, they spread out, so the models predict several 0.5$^\circ$ patches (cyan), one per group of RFs. Each patch is about the size of one LGN RF. This monkey saw the photo in a warmer color rendering. \textbf{B},~Spikes of these 152 units (rows) and gaze over time. Each jump in gaze is a saccade. Fixations are numbered as in A. The box marks fixation~1 (purple in A). \textbf{C},~The scene decoded by the diffusion model from a second monkey's LGN. The inset shows where LGN RFs landed over all fixations, trials and sessions.}
\newcommand{\CaptionSFour}{\textbf{Ridge-decoded patches and a scene prior for macaque LGN.} \textbf{A}, Measured gaze (yellow) and receptive-field patch locations (cyan; 2.2$^\circ$ from gaze). \textbf{B}, Spikes spanning four fixations, with decoded and true patches; dotted lines mark saccades and cyan each 100-ms decoder window, the one that overlaps each fixation most. \textbf{C}, Mosaics pool observations across sessions; the arrow follows patch~4 to its place. \textbf{D}, The prior denoises and fills gaps. Decoded contrast is shown $\times$3. Color is decoded in luminance and two opponent channels, with the prior scaled to each channel's power in the training photographs.}%

\newcommand{\CaptionThree}{\textbf{Ridge-decoded patches and a scene prior.} \textbf{A}, Measured gaze (yellow) and receptive-field patch locations (cyan). \textbf{B}, Spikes spanning four fixations, with decoded and true patches; dotted lines mark saccades, and cyan bands mark the 100-ms decoder windows. \textbf{C}, The mosaic is the weighted average of all decoded patches (Eq.~\ref{eq:mosaic}); the arrow places the decoded patch of fixation~4 from B at its position in C. \textbf{D}, The 1/$f$ prior denoises and fills in the gaps. Fig.~\ref{figS:lgn-ridge} shows the same for macaque LGN.}%
\newcommand{\CaptionFour}{\textbf{Transformer decoding and a learned diffusion prior.} The decoder, built from a transformer~\cite{vaswani2017} and Stable Diffusion~\cite{rombach2022}. Each token embeds a 100-ms window of spiking activity and the gaze track, at a 16.7-ms stride. A separate patch head trains the token embeddings to predict each window's local image patch. The transformer is trained from scratch; its decoded scene and the neural query features condition the diffusion model. Right: a held-out marmoset V1 scene, with its transformer and diffusion reconstructions.}
\newcommand{\CaptionMethods}{\textbf{Comparison with published brain decoders.} \textbf{A}, The table quantifies each method's correlation to the original photograph, overall and in different frequency bands. The transformer is the best in every column. \textbf{B}, Example reconstructions decoded from MEG, EEG, fMRI and electrophysiology. Methods are ordered by correlation to the original photograph. V1$\rightarrow$IT means V1, V4 and IT arrays. \textbf{C}, Perceptual similarity (LPIPS) versus correlation for selected decoders (all in Fig.~\ref{figS:landscape}); lines join our transformer (open) to diffusion (filled), and dashed steps mark the gaps between the best non-invasive, invasive and ex vivo decoders. Numbered as in B.}
\newcommand{\CaptionSBands}{\textbf{Reconstruction accuracy by spatial frequency, for every method.} Pearson correlation with the photograph in low (\mbox{0--0.49}), mid (\mbox{0.49--1.88}) and high (\mbox{1.88--4.88\,cycles/degree}) spatial-frequency bands and over the whole image, averaged over held-out images. Our monkeys show the transformer (open circles) and diffusion (filled). Thin stems show the patch mosaic and ridge regression with a 1/$f$ prior (Monkeys A and O). Published methods are as in Fig.~\ref{figS:landscape} (NSD: the 982 test images all four NSD methods share), grouped as in Fig.~5B. The retina's flashed images resolve only up to 3.1\,cycles/degree, so they have no high band (n/a).}
\newcommand{\CaptionTMetrics}{\textbf{Every decoder on standard reconstruction metrics.} Each method is scored with the same code on its own test images; Table~\ref{tabS:data} gives their number and the recordings, and Fig.~\ref{figS:landscape} and Appendix~\ref{app:comparison} the methods. Tiles are shaded from the worst to the best of all 30 decoders in each column, and discs mark the top three recording types, each by its best method. $r$ and LPIPS~\cite{zhang2018lpips} are the axes of Fig.~5C; the other metrics follow MindEye2~\cite{scotti2024mindeye2}, with PSNR and DINOv2 added. Two-way identification (\%, chance 50) uses AlexNet layers 2 and 5, Inception-v3, CLIP ViT-L/14~\cite{radford2021clip} and DINOv2 ViT-L/14 features; Eff (EfficientNet-B1) and SwAV are correlation distances.}
\newcommand{\CaptionTData}{\textbf{The data behind each decoder.} Rows are as in Table~\ref{tabS:metrics}; a value is printed where it first appears or changes within a recording type. $n$: test images. Train: training images from the decoded subject or animal (ours: per fold), marked $^\dagger$ where the decoder also trains on other subjects or animals (MindEye2 and Brain-IT on 60,566 images of the other NSD subjects, our model on all five monkeys); ---: not reported. Field: visual angle of the reconstructed image. Repeats: presentations of each test image whose responses the decoder uses (mean over images). Viewed: their total time on screen. Count: the units, cells, electrode sites, voxels or sensors the decoder reads. Sampling: how fast the recording system samples (fMRI: one volume per TR). Bin: the time step of the activity the decoder reads, or the window it averages into one value (fMRI: one response per 3-s presentation). Sources are in Appendix~\ref{app:comparison}.}
\newcommand{\CaptionSLandscape}{\textbf{Every published decoder and condition.} \textbf{A}, Fig.~5C with every published decoder and condition, numbered as in the key (N and F, the TVSD monkeys; 40\,h, 37\,h and 1\,h, fMRI data per subject), and all four of our monkeys; Table~\ref{tabS:metrics} scores the same methods on every standard metric. \textbf{B} (next page), Example reconstructions from every method, as in Fig.~5B (TVSD: monkey N). Each method is scored on its own test images, with the same code for every method. Brain-IT, MindEye2, Brain-Diffuser and Takagi: the 982 test images all four NSD methods share, mean of four subjects, with 40\,h or 1\,h of data per subject (Brain-Diffuser and Takagi: 37\,h); MindEye2 is scored from its released refined images, and Brain-Diffuser and Takagi from our runs of the released code (Takagi: its first sample). Fig.~5C scores Brain-IT on all 1,000 NSD test images; no NSD score differs between the two sets by more than .001. ATM, ENIGMA and Perceptogram: 200 THINGS-EEG2 images, mean of 10 subjects; ATM from its released images, ENIGMA and Perceptogram from our runs of the public code. ATM on MEG: 200 THINGS-MEG images, mean of four subjects, from our retrain of the public code. TVSD methods (MonkeySee, Perceptogram, FuzzyAlign and Ciferri et al.): 60 test images, with recordings from V1, V4 and IT (V1$\rightarrow$IT) of monkeys N and F; FuzzyAlign is our retrain of the public code. Brain2GAN: 50 face and 100 natural test images. Retina: 18 flashed and 16 jittered images selected by the authors, so these are not test-set averages. Our monkeys: 27 held-out photographs for Monkeys A and L, and 70 for Monkeys O and C. For our monkeys and the retina, LPIPS is computed on the largest square inside the covered region. The scores do not rank methods across datasets: the image sets differ, and our decoders cover only where the eyes went. Appendix~\ref{app:comparison} gives the methods and their sources.}
\newcommand{\CaptionSLGNViewing}{\textbf{Neural evidence, macaque LGN.} Fig.~6, repeated for one held-out macaque LGN scene. \textbf{A}, The scene after 10\,s to 10\,min of viewing, reconstructed by the four methods. The top row marks where receptive fields sampled the photograph. With 10 or 30\,s of evidence the diffusion model invents content (a tree trunk). The 10-min column is 580\,s, 79\% of all evidence. \textbf{B}, Accuracy over viewing time for the 70 held-out LGN scenes (69 at 10\,s). Left: the transformer's correlation with the photograph in three spatial-frequency bands, with 95\% bootstrap intervals. Middle and right: every decoder's correlation and LPIPS. \textbf{C}, The same scene as receptive field coverage increases across columns. Coverage is the share of the scene inside a sampled 0.5$^\circ$ patch, the patch that the transformer predicts in the LGN. The samples are densest around a simulated fovea (white cross). Each blue dot in the RF row is a sampled receptive field location, a 1$^\circ$ tile of receptive fields. The columns keep 2, 3, 5, 10 and 100\% of the evidence, so below 100\% only the 54\% of neurons assigned to tiles are used. The rows below set how many neurons of each session are used. \textbf{D}, Mean correlation with the photograph over the 70 scenes for C's grid, at ten levels of evidence. Contours show the coverage-neuron trade-off. From a fifth of the evidence on, coverage stays between 72 and 74\%, and more evidence adds samples at each location instead; those levels crowd the right edge. C and D use the transformer after fine-tuning with random neuron subsets.}
\newcommand{\CaptionSLGNGaze}{\textbf{Gaze accuracy, macaque LGN.} Fig.~7A, repeated for the macaque LGN. \textbf{A}, \textbf{B}, Two held-out scenes, reconstructed by the four methods with Gaussian gaze error added at test time. Columns give the standard deviation of the error. B shows the scene of Fig.~\ref{figS:lgn-viewing}. \textbf{C}, Correlation with the photograph in three spatial-frequency bands against gaze error, over the 70 held-out LGN scenes, with 95\% bootstrap intervals. The gray band marks 10--15$'$ of gaze error, as in Fig.~7C. This panel shows correlation where Fig.~7C shows signal retained: the LGN's high-frequency correlation is small even with exact gaze, so its retained share would be a ratio of two small numbers.}
\newcommand{\CaptionSFixation}{\textbf{More fixation examples.} Marmoset V1, as in Fig.~7B: five held-out images, including the two that Fig.~7B leaves out, reconstructed by the fixation transformer with Gaussian error added to every gaze coordinate at test time (standard deviation, in arcminutes), up to Fig.~7C's 20$'$. Correlation with the photograph at clean gaze, top to bottom: .26, .75, .26, .24 and .25.}
\newcommand{\CaptionFive}{\textbf{Neural evidence: viewing time, coverage and neuron count.} \textbf{A}, One held-out scene after 10\,s to 10\,min of viewing. \textbf{B}, Accuracy over viewing time for 27 scenes. Left: the transformer's correlation with the photograph in three spatial-frequency bands, with 95\% bootstrap intervals. Middle and right: every decoder's correlation and LPIPS. \textbf{C}, The same scene as receptive field coverage increases across columns. Each blue dot in the RF row is a sampled receptive field location. The rows below set the number of neurons at each blue dot. \textbf{D}, Mean correlation over the 27 scenes for C's grid. Contours show the coverage-neuron trade-off. Fig.~\ref{figS:lgn-viewing} shows macaque LGN.}

\newcommand{\CaptionSeven}{\textbf{Gaze accuracy limits spatial detail.} \textbf{A}, Free viewing: reconstructions with Gaussian gaze error added at test time; columns give its standard deviation. Fig.~\ref{figS:lgn-gaze} shows macaque LGN. \textbf{B}, Fixation: reconstructions by the fixation transformer with the same kind of gaze error, in arcminutes; Fig.~\ref{figS:fixation} shows more. \textbf{C}, Signal retained relative to clean gaze, in fixation (mid and high bands) and free viewing. Gray bands mark 10--15$'$ of gaze error, which leaves at most 16\% of the high-frequency signal in both fixation and free viewing.}

%% file: text/abstract.tex
Reconstructing images from brain activity has two broad goals. The first is to build brain--computer interfaces. The second is to understand how the brain sees. The eyes deliver incomplete, noisy input, so the brain must fill in the gaps with its expectations of what natural scenes look like. This filling-in relies on the brain's prior over natural images. Image reconstruction lets us test which natural image prior best matches the brain's natural image prior. However, most previous work on image reconstruction relies on an unnatural condition: subjects must hold their eyes still. In reality, primates constantly move their eyes to see the world. \textbf{We introduce a framework for reconstructing images from neural activity while a primate freely views a static natural scene.} We record from five monkeys with high-density electrode arrays in the lateral geniculate nucleus (LGN) and primary visual cortex (V1). The recorded receptive fields cover only a small part of the image. Thus, we combine neural activity across eye movements, trials, and sessions to reconstruct the full image. We first present a simple linear method with a 1/$f$ prior. We then scale this approach up with modern deep learning. We train a decoder by tokenizing spiking activity and measured gaze; these tokens are fed into a transformer and diffusion model to reconstruct the full scene. Three factors determine reconstruction quality: the amount of neural evidence combined over viewing time, the accuracy of eye tracking, and the choice of image prior. \textbf{Our method sets a new state-of-the-art performance in reconstructing natural images from brain activity, across brain recording modalities: fMRI, EEG, MEG and electrophysiology.}

%% file: text/introduction.tex
\paragraph{Brain Decoding for Visual Neuroscience:}
The neural coding problem can be broken down into the encoding and decoding directions.
Encoding uses the stimulus to predict neural activity, while decoding uses neural activity to reconstruct the stimulus~\cite{rieke1997spikes}.
The canonical relationship between encoding and decoding can be expressed with Bayes' rule, where $\mathbf{s}$ is the stimulus and $\mathbf{r}$ is the neural response:
\begin{equation}
p(\mathbf{s} \given \mathbf{r}) = \frac{p(\mathbf{r} \given \mathbf{s})\, p(\mathbf{s})}{p(\mathbf{r})}
\label{eq:bayes}
\end{equation}
The neuroscience community often studies the encoding direction or $p(\mathbf{r} \given \mathbf{s})$ in order to test how well we can model the computation of neurons.
An encoding model that generalizes to any class of stimuli would be a true ``digital twin'' of the neurons.
However, the decoding direction can also illuminate interesting properties of the brain.
For example, work in macaque inferotemporal cortex (IT) showed that a linear decoder reading just ${\sim}200$ face-selective neurons can reconstruct a face, once faces are described in a 50-dimensional space of shape and appearance~\cite{chang2017faceid}.
This revealed a simple code for facial identity: no neuron stands for a particular face; each neuron measures one feature of a face, and together the population of neurons describes any face.
But faces are a special case.
Natural scenes do not live in a 50-dimensional space, so decoding them depends on choosing the right image prior.

In this paper, we study the decoding problem through the lens of reconstructing natural images from neural activity.
We are building a model for $p(\mathbf{s} \given \mathbf{r})$, where the stimulus $\mathbf{s}$ is the image shown.
In order to build a model for $p(\mathbf{s} \given \mathbf{r})$, we have to make explicit choices about our prior of natural images, $p(\mathbf{s})$.
This provides a way to test our models of natural image priors and assess their consistency with the information represented in the brain.
This may eventually help answer a long-standing question in visual neuroscience: \textbf{What is the brain's natural image prior model?}

\newpage
\paragraph{Brain Decoding for Brain Computer Interfaces (BCIs):}
Vision restoration efforts are moving towards invasive brain implants.
Neuralink's Project Blindsight~\cite{neuralink2026visual} will implant electrodes in V1 and Phosphoenix~\cite{phosphoenix2026financing} in the LGN, the same two areas we decode from.
However, these current implants are only able to record from a tiny fraction of the full population of neurons.
The human visual cortex contains 4 to 6 billion neurons~\cite{wandell2009visual}, but current BCIs can only record from hundreds of neurons at most.
To make matters worse, these visual neurons usually only represent a very tiny localized region of the full visual field.
Given these heavy constraints, we ask the pragmatic question: \textbf{How can BCIs operate from a severely subsampled population of neurons?}

\paragraph{Contributions:}
This work is motivated by two lofty goals: discovering the natural image prior in the brain and building BCIs for vision restoration.
This work does not solve either problem, but it is a necessary first step.
Previous work in primates has reconstructed images only while the monkey holds its gaze still, never during free viewing of natural images.
We make the following contributions:
\begin{enumerate}[label=\arabic*),leftmargin=*,nosep,topsep=2pt]
\item \textbf{The first method for reconstructing natural images during free viewing} from V1 and LGN (Figs.~3 and~4).   %
For LGN, the first \textit{color} image reconstructions and the first reconstructions from primates.
\item \textbf{A comparison of decoding performance across brain recording modalities}: fMRI, EEG, MEG, electrophysiology (Fig.~5).
State-of-the-art reconstruction in awake primates (humans and monkeys).
\item \textbf{A quantification of reconstruction quality across image priors}: no prior, 1/$f$, deep network's learned prior, diffusion model's generative prior (Figs.~1 and~5).
\item \textbf{A measurement of the neural evidence needed for faithful reconstructions}: total viewing time, neuron count, and receptive field coverage (Fig.~6).
\item \textbf{Evidence that eye tracking is essential}: both tiny and large eye movements must be tracked (Fig.~7).
\item \textbf{Public release of V1 data, model, and code}.   %
\end{enumerate}

%% file: text/related_work.tex
\paragraph{History of Reconstructing Images from Brain Activity:}
Reconstructing images from brain activity largely falls into two recording modalities: functional magnetic resonance imaging (fMRI) and electrophysiology.
The first demonstration of reconstructing naturalistic images from recorded spike trains was from the LGN of anesthetized, paralyzed cats in 1999~\cite{stanley1999lgn}.
In 2009, naturalistic images were reconstructed from fMRI in humans~\cite{naselaris2009bayesian}, followed by natural movies in 2011~\cite{nishimoto2011reconstructing}.
Eye movements were minimized in all of these studies.
These methods used simple machine learning: a linear decoder for the LGN, and for fMRI, linear encoding models used to pick the best-matching images or movie clips from a database.
After the invention of diffusion models in 2015~\cite{sohldickstein2015deep}, methods in both fMRI and electrophysiology have moved to diffusion, with the first demonstrations from human fMRI~\cite{takagi2023high,chen2023seeing,allen2022massive}.
While improvements have been made to fMRI diffusion decoding~\cite{beliy2026brainit}, reconstructions from fMRI largely produce plausible images rather than faithful reconstructions of the shown stimuli~\cite{shirakawa2025spurious}.
Part of the reason is that fMRI measures blood oxygen level, not neural activity; it can cover the entire brain, but each voxel pools hundreds of thousands of neurons.\looseness=-1 %

\paragraph{Reconstructing Images from Electrophysiological Recordings:}
On the other hand, electrophysiological recordings from non-human primates carry far more precise information about each individual neuron's activity, allowing for more faithful reconstructions.
Several prior methods have reconstructed images from ventral stream recordings.
The ventral stream, which carries the `what' of the scene, runs from the retina $\rightarrow$ LGN $\rightarrow$ V1 $\rightarrow$ V2 $\rightarrow$ V4 $\rightarrow$ IT.
Roughly speaking, the retina can be thought of as the camera sensors of the brain, V1 is an edge detector, V2 is selective for contours and textures, V4 is selective for curved shapes and color, and IT is known for face and object selectivity.
In the primate visual system, natural images have been reconstructed from ex vivo retinas (living tissue separated from the animal)~\cite{wu2022map,brackbill2020reconstruction,kim2021nonlinear}; V1, V4 and IT responses to flashed THINGS images (TVSD)~\cite{papale2025tvsd,le2024,le2025irfa,fei2025perceptogram,ciferri2026modular,song2026fuzzyalign}; GAN-generated images~\cite{dado2024brain2gan}; faces and objects in IT~\cite{hayashi2018gan,chang2017faceid,bao2020objectspace,hesse2026rapid}; dorsal-stream video~\cite{trepka2025stsbench}; task videos across cortex~\cite{zhang2026anndecoding}; and anesthetized V1~\cite{li2023braininspired}.
\textbf{No prior work has reconstructed images from \textit{primate} LGN.}

\paragraph{Prior Work Lacks Eye Movements:}
Reconstruction methods based on fMRI or electrophysiology have not taken eye movements into account.
However, during natural conditions, the eyes move around the scene, largely due to saccades which occur 2--3 times a second.
Neural activity in V1, V2, V4, and IT has been shown to be sensitive to eye movements~\cite{xiao2024feature}.
More recent advances in eye tracking and free-viewing setups~\cite{yates2023,ressmeyer2026} have shown that V1 neural activity is sensitive to the tiny eye movements that happen during fixation~\cite{ressmeyer2026active}.
Theoretical models of reconstruction during fixational eye movements demonstrate the importance of accounting for these tiny shifts in the image and the associated neural activity~\cite{anderson2020highacuity,burak2010bayesian,pitkow2007neural,rucci2007miniature}.
The only previous experimental work that accounts for eye movements reconstructs images from an ex vivo retina with simulated fixational eye movements~\cite{wu2024fixational}.
This work shows that accounting for fixational eye movements improves reconstruction quality.
\textbf{No prior work has reconstructed images from neural activity of awake, behaving, freely viewing primates.}

\paragraph{Reconstructing Images for BCIs:}
A natural question is: what is the point of reconstructing images for vision restoration?
Conventionally, vision restoration is understood as an electrical stimulation problem, not a decoding problem~\cite{holz2026prima,beauchamp2020dynamic}.
However, reconstruction and stimulation will be equally important: the image reconstructed from the evoked neural activity should match the image the implant intended.
Implants that both stimulate and record now make this check possible, first in the ex vivo primate retina~\cite{shah2024precise,phillips2026electrical} and recently in the visual cortex of a blind person~\cite{moure2026deep}.\looseness=-1

%% file: text/fig5_refs.tex
\FigRefAt{0.13181}{0.30265}{li2024atm}
\FigRefAt{0.28583}{0.30265}{fei2024perceptogram}
\FigRefAt{0.43986}{0.30265}{beliy2026brainit}
\FigRefAt{0.59389}{0.30265}{dado2024brain2gan}
\FigRefAt{0.13181}{0.01212}{le2024}
\FigRefAt{0.20359}{0.01212}{fei2025perceptogram}
\FigRefAt{0.35761}{0.01212}{dado2024brain2gan}
\FigRefAt{0.96324}{0.01212}{wu2024fixational}

%% file: text/methods.tex
\paragraph{Recording Setup:}
We decoded recordings from five monkeys during natural vision.
Two marmosets (Monkeys A and L) freely viewed the same 27 \textit{grayscale} photographs while we recorded from V1 with multisite silicon arrays~\cite{yates2023}; a third (Monkey L2) was used for training only.
Two macaques (Monkeys O and C) viewed 70 \textit{color} photographs while we recorded from the LGN with Neuropixels probes.
In every LGN session, small Gabor targets appeared on the photograph, and the monkey fixated them to receive a reward.
Every photograph was shown for 10\,s.
Summed over all sessions, each photograph was viewed for about 7\,min on average.
Eye position was measured with an eye tracker~\cite{wu2023dpi,ressmeyer2026}.
All decoders used spikes and eye position binned at 120\,Hz (8.3-ms bins).
In total, we used 94 sessions and 23,452 units, about 250 units per session.

\paragraph{Problem Statement:}
Figs.~2 and~\ref{figS:lgn-decoding} illustrate the problem for V1 and LGN, respectively.
The goal is to reconstruct the full image across sessions, trials and eye movements.
Each fixation while the monkey is freely viewing the scene is a snapshot of evidence to reconstruct it.
Importantly, the receptive fields of the neurons are fixed relative to the center of gaze and cover only a tiny fraction of the image at any one fixation.
However, across many fixations, the receptive fields eventually cover most of the image.
Because we track gaze, we know where these receptive fields land at any moment: always a constant translation away from gaze.
We reconstruct the region of the image where there is an appropriate amount of evidence.
All decoders follow the same principle: 1)~decode local patch-level information from the spikes at each recorded gaze position; 2)~combine the local evidence into a full scene; and 3)~fill in the gaps where the receptive fields didn't land using an image prior.

\newpage
\subsection{Ridge Regression to Decode Image Patches}
\label{sec:ridge}
We first decode the small patch of the photograph at the receptive fields of the recorded neurons, one short window of spikes at a time (Fig.~3).
Because we track gaze, we know which patch fell on the receptive fields in every window, so every window is a training pair: spike counts in, image patch out.
We cut each recording into back-to-back 100-ms windows, take the median gaze in each, and count spikes over the same 100\,ms, shifted 16.7\,ms later to allow for the response delay.
In V1, the patch is $21\times21$ grayscale pixels (0.56$^\circ$), centered on the population receptive field.
In LGN, we decode a color patch ($13\times13$ samples over 0.75$^\circ$, in luminance and two opponent channels) for each group of neurons with nearby receptive fields.

\textbf{Predicting patches from spikes} (Fig.~3B): Ridge regression predicts from a window's spike counts $\mathbf r$ (log-transformed and standardized) how its patch $\mathbf y$ differs from the average training patch $\bar{\mathbf y}$:
\begin{equation}
\hat{\mathbf d}=W\mathbf r,\quad W=\arg\min_W\sum_n\lVert\mathbf d_n-W\mathbf r_n\rVert^2+\lambda\lVert W\rVert^2,
\label{eq:ridge}
\end{equation}
with $\mathbf d_n=\mathbf y_n-\bar{\mathbf y}$ for training window $n$.
A different group of neurons is recorded in each session, so we fit one decoder per session (and per LGN group).
We choose $\lambda$ by cross-validation on training photographs.

Spikes are noisy, so this decoder hedges: on average, $\hat{\mathbf d}$ is a shrunken version of the true $\mathbf d$, and decoded patches look washed out (Fig.~3B).
So we first scale $\hat{\mathbf d}$ back to full strength and add back the average patch:
\begin{equation}
\mathbf p=\bar{\mathbf y}+\hat d_{\mathrm{bright}}/g_{\mathrm{bright}}+\hat{\mathbf d}_{\mathrm{contrast}}/g_{\mathrm{contrast}}.
\label{eq:gains}
\end{equation}
The mean of $\hat{\mathbf d}$ over its pixels is a scalar, $\hat d_{\mathrm{bright}}$, which says how much brighter or darker the whole patch is than the average training patch $\bar{\mathbf y}$.
Subtracting $\hat d_{\mathrm{bright}}$ from every pixel of $\hat{\mathbf d}$ leaves a vector, $\hat{\mathbf d}_{\mathrm{contrast}}$, the pattern of light and dark inside the patch.
Both $\hat d_{\mathrm{bright}}$ and $\hat{\mathbf d}_{\mathrm{contrast}}$ have their own gain, $g=\langle\hat d\,d\rangle/\langle d^2\rangle$: the fraction of the true signal that the decoder recovers on training photographs held out from the ridge fit.
On these held-out photographs, we also measure each decoder's reliability $\pi$: the inverse of its mean squared error after calibration. Each decoder's $\pi$ roughly estimates how reliable that session's patch predictions are.

\textbf{Placing patches using measured gaze} (Fig.~3C): Gaze places each calibrated patch in the scene, at the gaze position plus the receptive-field offset.
A scene thus collects $K$ patches, $\mathbf p_1,\dots,\mathbf p_K$, one from every spike window of every trial and session; each patch pixel is a noisy measurement of the photograph.
At each pixel location $\mathbf x$ of the full scene, the patch mosaic $M(\mathbf x)$ is the average of all the patch pixels $p_k(\mathbf x)$ that land on $\mathbf x$ (from patches $k\ni\mathbf x$), each weighted by its decoder's reliability $\pi_k$.
The mosaic uses no image prior, so it is noisy and full of gaps.
The sum of the reliabilities at $\mathbf x$ is the overall reliability, $w(\mathbf x)$:
\begin{equation}
M(\mathbf x)=\sum\nolimits_{k\ni\mathbf x}\pi_k\,p_k(\mathbf x)\,\big/\,w(\mathbf x),\quad w(\mathbf x)=\sum\nolimits_{k\ni\mathbf x}\pi_k.
\label{eq:mosaic}
\end{equation}

\subsection{Patch Mosaic + \texorpdfstring{$1/f$}{1/f} Prior}
\label{sec:prior}
\textbf{Denoising and filling in with 1/$f$ prior} (Fig.~3D): We denoise and fill in the mosaic $M$ with the simplest model of natural images, a $1/f$ prior~\cite{field1987}.
In natural images, low spatial frequencies have more power than high ones.
We solve for the image $\hat I$ that fits the measurements and is likely under the prior:
\begin{equation}
\hat I=\arg\min_I\sum_{\mathbf x}w(\mathbf x)\bigl(I(\mathbf x)-M(\mathbf x)\bigr)^2+\gamma\sum_{\mathbf f}\lvert\mathbf f\rvert^{2}\,\lvert\tilde I(\mathbf f)\rvert^2.
\label{eq:map}
\end{equation}
The first term keeps the image close to the mosaic, weighting each pixel by its reliability, $w(\mathbf x)$; in the gaps, $w(\mathbf x)=0$.
The second is the prior on $\tilde I$, the Fourier transform of the candidate image: higher spatial frequencies are penalized more heavily than lower ones.
With $\gamma=0$, the solution is the mosaic itself; as $\gamma$ grows, noise is smoothed away and the gaps are filled (Fig.~3D).
We choose $\gamma$ on training photographs.
We solve Equation~\ref{eq:map} with conjugate gradients, a standard iterative solver (see Appendix~\ref{app:derivation}).

\newpage
\subsection{Transformer Decoder}
\label{sec:transformer}
Ridge regression decodes each window on its own; a fixed 1/$f$ prior is applied afterwards.
The transformer instead reads all the spike windows of a scene together and learns its prior from the training photographs (Fig.~4; Appendix~\ref{app:offset}).

\textbf{Tokens.}
We cut each recording into 100-ms windows with a 16.7-ms stride.
Each window becomes a 256-dimensional token, through a read-in layer for its session (every session records different neurons).
The token also encodes gaze, which places it where the receptive fields landed.
Each photograph, viewed for over 10\,min, gives about 41,000 tokens (Monkey A) or 30,000 (Monkey O).

\textbf{Self-attention}~\cite{vaswani2017} lets every token exchange information with every other, so each window is read in the context of the whole recording.
For the \textbf{global context}, 256 learned latents read all the tokens and summarize the scene~\cite{jaegle2021perceiver}.

\textbf{Query readout.}
A $45\times80$ grid of queries covers the image, one per $4\times4$-pixel block.
Each query reads its \textbf{local context}, the 512 tokens closest to it, which carry what fell on the receptive fields around its block; it alternates this with reading the global context, for the rest of the scene.
A small convolutional network then smooths the seams, giving a $180\times320$ scene: grayscale for V1 and color for LGN.

\textbf{Training.}
We train one model from scratch on data from all five monkeys, with five folds split by photograph.
We use a robust L1 (Charbonnier) loss against the photograph on the pixels the receptive fields covered, plus 2.5 times the same loss on high-pass-filtered images, which rewards fine detail.
A \textbf{patch head} also decodes each token to predict the image patch where the receptive fields landed during that token's 100-ms spike window.
Two augmentations help the model generalize from as few as 27 photographs.
First, each training example uses a random subset of a photograph's tokens, so the model learns to decode from few or many tokens.
Second, each example shifts the photograph by a random offset and shifts all gaze positions by the same offset.
At test time, the model reads every token of a held-out photograph.

\subsection{Stable Diffusion Prior}
\label{sec:diffusion}
The transformer-decoded scenes are faithful but blurry.
To add realistic detail, we condition a pretrained image generator on them (Fig.~4).
We start from MindEye2's~\cite{scotti2024mindeye2} unCLIP version of Stable Diffusion XL (SDXL)~\cite{podell2024sdxl}, which turns a CLIP image embedding~\cite{radford2021clip} back into an image.
We freeze the transformer and give the generator two inputs.
The first is the CLIP embedding of the decoded scene, refined by an adapter that is trained to match the photograph's CLIP embedding.
The second is the transformer's final $45\times80$ query features, which another adapter feeds into the generator's denoising network to tell it where things are.
In addition to training these two adapters, we fine-tune the generator's attention layers with LoRA~\cite{hu2022lora}.
A standard denoising loss is used.
In training, the scenes are decoded from random subsets of each photograph's tokens.
Generation starts from random noise, so the decoded scene is a condition, not a starting image.
We draw one sample per scene, with a fixed seed.

%% file: text/results.tex
\section{Results}\label{sec:results}
\subsection{Comparison of Decoding Methods}
\paragraph{Comparison of Image Priors:}
The patch mosaic + 1/$f$ decoder serves as a simple, yet powerful proof of concept.
However, the patch mosaic + 1/$f$ method does not create realistic images.
Going from the simple patch mosaic + 1/$f$ to the transformer increases the parameter count from 1M to 17M.
In exchange for the increase in parameters, the transformer reconstructions are better, especially in V1 at mid and high spatial frequencies, which are necessary to make images look sharp and structured (Fig.~5A).
Diffusion grows to 4.4 billion parameters but makes the reconstructions perceptually much closer to the original photographs (Fig.~5C).
Perceptual similarity is measured with Learned Perceptual Image Patch Similarity (LPIPS)~\cite{zhang2018lpips}.
However, diffusion does worse than the transformer on every correlation measure, which demonstrates the fidelity--realism trade-off~\cite{blau2018perception}.

\newpage
\paragraph{Comparison to Previous Brain Decoders:}
We placed the mainstream decoding approaches for each brain recording modality by correlation with the photograph and perceptual similarity (LPIPS; Fig.~5C).
\textbf{Our method is the state of the art for reconstructing natural images from awake behaving primates, whether humans or monkeys.}
Two disclaimers are important.
First, each prior work is scored only on its own dataset (Table~\ref{tabS:data}); there is no cross-dataset comparison.
Second, the amount of neural evidence behind each reconstruction is not represented in Fig.~5C.
fMRI records from the whole brain, but its time resolution is slow and each voxel pools hundreds of thousands of neurons.
We record from a few hundred neurons, but over ten minutes per image; this variable is hard to pin down on one scale.

The goal of this work is not to set up a classic machine learning benchmark problem.
Instead, it is to show the gaps between related methods: the best non-invasive, invasive and ex vivo methods (dashed steps in Fig.~5C).
Technically, Brain2GAN on faces (7) has the best perceptual similarity of any invasive decoder in awake monkeys.
However, its stimuli are faces, which live in a roughly 50-dimensional space~\cite{chang2017faceid} and are far easier to decode than natural scenes; on natural images from the same monkey, its performance is lower (4).
We also believe that state-of-the-art reconstruction here is not a result of our machine learning.
Rather, it is a result of the engineering that went into collecting the data: precise eye tracking and high-density neural recordings.
In Fig.~5B, ordered by correlation, many methods, especially diffusion-based ones, produce plausible but unfaithful images.
Table~\ref{tabS:metrics} adds more metrics: fMRI leads on two high-level ones, CLIP and DINOv2.
Ours has the best combination of correlation and perceptual similarity because it combines evidence over eye movements.
This is evidence that recording during free viewing, with eye movements measured, beats fixation: the receptive fields move with the eyes, so each eye movement brings new evidence.

\subsection{Neural Evidence}
\paragraph{Viewing Time:}
The best reconstructions come from combining evidence over long viewing times (Fig.~6A).
More viewing time means the receptive fields of the recorded neurons cover more of the scene, providing more neural evidence for the reconstruction.
The benefit of increased viewing time is spatial frequency-dependent.
As viewing time increases, the correlation between the original image and the reconstruction saturates quickly at low spatial frequencies, but it continues to increase at mid- and high spatial frequencies (Fig.~6B).

\paragraph{Receptive Field Coverage and Neuron Count:}
To mimic the coverage that an implant such as Neuralink's might have, we use foveated sampling~\cite{fritsches1996,rosa2000}: dense at the fovea and sparse in the periphery.
We vary how much of the scene is covered by receptive fields and how many neurons contribute at each location (Fig.~6C).
Coverage and neuron count trade off (Fig.~6D).
With 16\% coverage and $\approx$160 neurons contributing to each location, reconstructions match those from 20 neurons per location at 66\% coverage (.67 vs.\ .69).
If only 5 neurons contribute to each location, the correlation never exceeds .59.

\newpage
\subsection{Eye Movements}
Adding error to the measured gaze quickly degrades free-viewing reconstructions (Fig.~7A).
Detail is lost from fine to coarse: an error of 0.2$^\circ$ (12$'$) leaves 16\% of the high-frequency signal, 0.5$^\circ$ leaves 26\% of the mid band, and the low band still keeps 76\% at 1$^\circ$ (Fig.~7C).
The overall correlation hides this loss: at 1$^\circ$ it falls only from .80 to .70.
We also decode during fixation~\cite{ressmeyer2026active}.
The marmosets fixate a point while a repeated sequence of whitened natural images, each 3$^\circ$ across, flashes at 20\,Hz (Appendix~\ref{app:rec-al}).
Even then, a gaze error of just 5 arcminutes degrades the image substantially (Fig.~7B): the high-frequency signal falls to 62\%, and to 2\% at 15$'$.
\textbf{Fixation and free viewing agree (Fig.~7C): high-frequency information drops off very quickly when fixational eye movements are not measured.}
In fixation and in free viewing, a gaze error of about 10$'$ leaves only a sixth of the high-frequency signal.
This matters for BCIs, which would track gaze with wearable eye trackers that currently report a median error of 1.3--1.8$^\circ$ (78--108$'$)~\cite{baumann2026neon}.
At that error, only the coarse layout of a scene survives: the mid and high bands keep about 8\% of their signal or less.

%% file: text/discussion.tex
\newpage
\section{Discussion}\label{sec:discussion}
\paragraph{A New Baseline for Brain Decoding:}
Nearly three decades after natural scenes were first reconstructed from the LGN of anesthetized cats~\cite{stanley1999lgn}, this work marks how far brain decoding has come.
This progress rests on three important components: 1)~high-density neural recordings~\cite{jun2017neuropixels,yates2023}, 2)~accurate and precise eye tracking~\cite{yates2023,ressmeyer2026}, and 3)~strong image priors learned by transformers~\cite{vaswani2017} and diffusion models~\cite{sohldickstein2015deep}.
Without enough neurons (Fig.~6C) or accurate eye tracking (Fig.~7), reconstruction struggles.
Without strong image priors, reconstructions are not realistic (Figs.~5C and~6B).

\paragraph{Image Priors:}
Ridge regression with a 1/$f$ prior shows that much of the image content is linearly available in the spikes, with a simple prior filling the gaps.
Similar to face decoding from IT~\cite{chang2017faceid}, we demonstrate that no black box is needed to decode natural images during free-viewing.
We suspect the transformer learns a stronger image prior.
The diffusion model produces realistic images because it samples them from a prior learned on a large corpus of natural images~\cite{sohldickstein2015deep}.
Sampling makes a diffusion model creative~\cite{kamb2025creativity}: where the neural evidence runs out, it invents.
While this work does not answer what the natural image prior of the brain is, it is an exercise in studying state-of-the-art image priors and seeing where they fall short.
The ideal image prior should be able to take partial observations of brain activity and recreate the subject's full visual perception.
This ideal model would have to capture everything from edge detectors in V1~\cite{olshausen1996sparse} to how we perceive faces in IT~\cite{chang2017faceid} to the way illusions trick our visual system~\cite{weiss2002motion}.
We release the V1 dataset, model weights, and code so that others can test their own image priors and models of the brain.

\pagebreak
\paragraph{Brain Decoding Across Recording Modalities:}
Image reconstruction makes a practical benchmark for visual neuroscience.
Encoding benchmarks report neural prediction with fraction of variance explained, correlation, $R^2$, or bits/spike~\cite{schrimpf2018brainscore,willeke2022sensorium,pei2021nlb}.
What counts as a good score depends on each dataset's noise ceiling, so these numbers are difficult to interpret and do not generalize across recording modalities.
A reconstruction puts every recording in the same space---the image.
The comparison across recording modalities in this paper (Fig.~5) would be far harder to understand with encoding models.

This work is the first to quantify the gap between non-invasive and invasive decoding (Fig.~5C), and we hope future work closes that gap, especially for non-invasive BCIs.
We also show how far the best awake recordings are from the best ex vivo ones.
\textbf{Our immediate proposal to all brain decoding efforts across recording modalities is to track eye movements.}
In fMRI, gaze can even be estimated from the position of the eyes in the functional images~\cite{frey2021deepmreye}, so existing datasets could be reconstructed with an explicit gaze signal.

\paragraph{BCIs for Vision Restoration:}
Vision restoration for people with partial vision, such as retinitis pigmentosa or macular degeneration, will require a BCI that decodes from undamaged representations of visual field locations and writes into representations that are damaged or no longer receive signal.
\textbf{We believe that BCIs for vision restoration should combine electrical stimulation with online brain decoding.}
Based on this work, we propose three components: (1)~accurate gaze tracking, with an online version of neural shifter calibration~\cite{yates2023}; (2)~an online decoder like ours; and (3)~broad receptive-field coverage.

%% file: text/supp_nogaze.tex
\newcommand{\CaptionTNoGaze}{\textbf{Decoding without eye position.} Pearson correlation with the photograph over the whole image and in the low, mid and high spatial-frequency bands of Fig.~\ref{figS:bands}. Values are means over held-out photographs: 27 for Monkeys A and L, 70 for Monkeys O and C. \emph{Spikes and eye position} is the transformer of the paper. \emph{Spikes only} is the transformer retrained without eye position. \emph{Eye position only} and \emph{Neither} are these two models with every spike count set to zero at test time.}
\newcommand{\CaptionSNoGaze}{\textbf{Reconstructions without eye position.} Five held-out scenes for each area, drawn as in Figs.~\ref{figS:v1-scenes} and~\ref{figS:lgn-scenes}. The top three of each area are those of Fig.~1's grid. The four columns on the right are the four rows of Table~\ref{tabS:nogaze}. With spikes only, the transformer paints a picture close to a training photograph. With neither input, it paints the same few pictures whatever the scene.}
\section{Decoding without Eye Position}\label{app:nogaze}Does the transformer need to know where the eye is?
We retrained the transformer of Section~\ref{sec:transformer} without eye position.
Every token is placed as if the eye rested at the centre of the display.
Each group of receptive fields keeps only its fixed offset from the eye.
For Monkeys A and L, as in the transformer of the paper, a session's neurons form one group with no offset, so every one of their tokens sits at the centre.
The model is never told where the eye is.
Two further inputs, the eye's trajectory within each spike window and the decoder's map of where the eye went, are held constant.
The patch head is not trained, because its targets are cut from the photograph at the gaze position.
The spike windows, the architecture, the five training monkeys, the five folds and the training schedule are unchanged.

\textbf{Without eye position, the reconstruction is lost.}
Image $r$ falls from .80 to .29 for Monkey A, from .66 to .27 for Monkey L, from .68 to .32 for Monkey O and from .57 to .27 for Monkey C (Table~\ref{tabS:nogaze}).
The mid and high bands fall below .06.
Image $r$ is lower in each of the 27, 27, 70 and 70 photographs.
What the retrained transformer paints is close to a photograph from its training set (Fig.~\ref{figS:nogaze}).
Its output correlates with the closest training photograph at a median of .76, .83, .51 and .51 for the four monkeys, and with the photograph shown at .32, .23, .29 and .23.
The transformer of the paper shows the reverse: .50, .52, .40 and .39 with the closest training photograph, and .83, .66, .72 and .56 with the photograph shown.
For reference, a photograph correlates with its own closest training photograph at .46, .39, .34 and .33.
The Image $r$ that remains therefore measures how much the recalled photograph resembles the photograph shown.

\textbf{Neither input is enough alone.}
Setting every spike count to zero at test time gives the other two rows of Table~\ref{tabS:nogaze}.
With eye position alone, the transformer of the paper scores an Image $r$ of .21, .27, .01 and $-$.05.
With neither input, the retrained transformer scores .24, .21, .04 and $-$.02.
In V1, spikes alone add little to that, .05 for Monkey A and .06 for Monkey L.
In LGN, spikes alone add .28 and .29, almost all in the low band.
LGN spikes steer the output toward a training photograph more like the photograph shown: the closest training photograph correlates with the photograph shown at a median of .29 and .20, against .10 and .02 with the spikes set to zero.
Only spikes and eye position together reconstruct the scene.

\textbf{What still depends on eye tracking.}
The control uses the same spike windows as the transformer of the paper: the windows with valid tracking while the animal viewed the photograph.
The model still reads flags that mark when tracking was valid, and groups without valid tracking are masked.
On the held-out scenes of one fold, no group was masked because of where the eye was.
The training loss and the scores cover the pixels that the receptive fields covered, as in the paper.
In LGN, each group's fixed offset from the eye is its median over the session's windows with valid tracking.
For Monkey L2, the monkey used only for training, groups whose receptive fields fell off the photograph are masked during training.

{\centering\small\setlength{\defaultaddspace}{2pt}\setlength{\abovecaptionskip}{5pt}
\captionof{table}{\CaptionTNoGaze}
\label{tabS:nogaze}\nopagebreak
\input{text/table_nogaze}\par}
\clearpage
\noindent\makebox[\linewidth][c]{\includegraphics{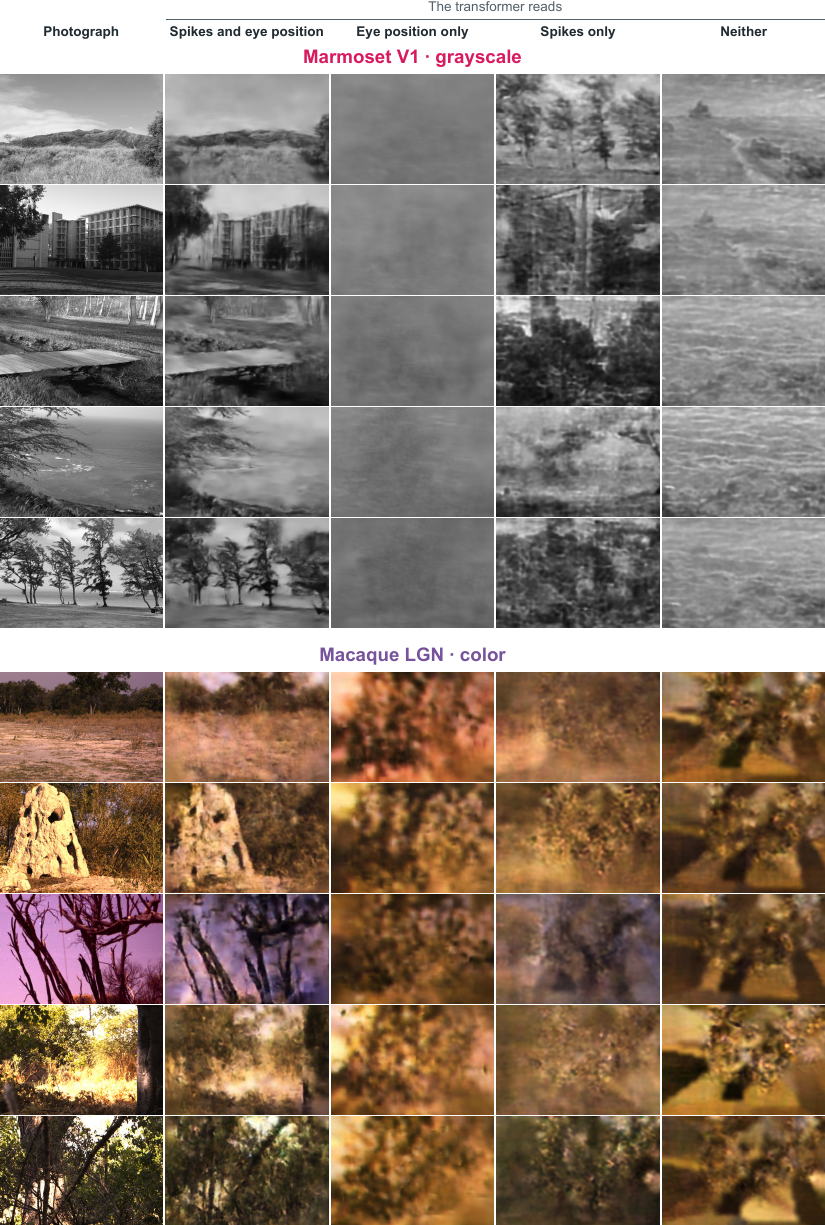}}\par{\captionof{figure}{\CaptionSNoGaze\label{figS:nogaze}}}

%% file: text/table_nogaze.tex
\begin{tabular}{@{}llcccc@{}}
\toprule
Monkey & The transformer reads & Image $r$ & Low & Mid & High \\
\midrule
A (V1) & Spikes and eye position & .803 & .902 & .582 & .252 \\
 & Eye position only & .210 & .236 & .039 & .011 \\
 & Spikes only & .289 & .349 & .013 & .018 \\
 & Neither & .244 & .301 & .012 & .006 \\
\addlinespace
L (V1) & Spikes and eye position & .656 & .765 & .324 & .115 \\
 & Eye position only & .269 & .301 & .073 & .025 \\
 & Spikes only & .271 & .320 & .025 & .015 \\
 & Neither & .210 & .261 & .016 & .005 \\
\addlinespace
O (LGN) & Spikes and eye position & .678 & .814 & .487 & .117 \\
 & Eye position only & .012 & .010 & .006 & .001 \\
 & Spikes only & .316 & .378 & .046 & .038 \\
 & Neither & .038 & .043 & .007 & .010 \\
\addlinespace
C (LGN) & Spikes and eye position & .569 & .707 & .311 & .061 \\
 & Eye position only & \llap{$-$}.049 & \llap{$-$}.063 & \llap{$-$}.004 & \llap{$-$}.003 \\
 & Spikes only & .275 & .324 & .054 & .039 \\
 & Neither & \llap{$-$}.020 & \llap{$-$}.026 & .008 & .013 \\
\bottomrule
\end{tabular}

%% file: text/appendix_comparison.tex
\section{Comparison with Published Decoders}
\label{app:comparison}
Figure~5 places our decoders beside published reconstructions from other recordings.
For human fMRI, we use Brain-IT~\cite{beliy2026brainit} and MindEye2~\cite{scotti2024mindeye2} on the Natural Scenes Dataset~\cite{allen2022massive}, trained on 40\,h or 1\,h of data per subject.
The supplement (Figs.~\ref{figS:landscape} and~\ref{figS:bands}) adds Brain-Diffuser~\cite{ozcelik2023natural} and Takagi and Nishimoto~\cite{takagi2023high}, from our runs of the released code, which trains on each subject's first 37 sessions (37\,h); for Takagi, we score its first sample.
There, all four NSD methods are scored on the 982 test images they share; Figure~5C scores Brain-IT on all 1,000.
For human EEG, we use ATM~\cite{li2024atm}, ENIGMA~\cite{kneeland2026enigma} and Perceptogram~\cite{fei2024perceptogram} on THINGS-EEG2~\cite{gifford2022things}.
For ATM, we use the released reconstructions.
For ENIGMA, we retrain the single-subject model from its public code on ATM's preprocessed EEG, because ENIGMA's own preprocessing is not released; our retrain scores up to 2.6 points below the published results.
For Perceptogram, we run its public code.
Like Perceptogram on TVSD~\cite{fei2025perceptogram}, it standardizes its predicted latents by the test predictions' own mean and standard deviation, a step that uses all 200 test images together.
For human MEG, we retrain ATM from its public code on THINGS-MEG~\cite{hebart2023things}, with minimal fixes; the retrain reproduces the paper's MEG results within margins we set in advance.
We leave out Perceptogram on MEG, because its released code cannot reproduce the paper's MEG result.
For macaque electrophysiology, we use MonkeySee~\cite{le2024}, Perceptogram~\cite{fei2025perceptogram}, FuzzyAlign~\cite{song2026fuzzyalign} and Ciferri et al.~\cite{ciferri2026modular} on TVSD~\cite{papale2025tvsd}, for monkeys N and F, and Brain2GAN~\cite{dado2024brain2gan} on faces and on natural images.
For the retina, we use LNBRC-dCNN by Wu et al.~\cite{wu2024fixational}, for flashed and jittered images; only the authors' selected example images are public, so these scores are not averages over a test set.

Every method is scored with the same code.
Table~\ref{tabS:metrics} gives every method's scores on the metrics most decoders report.
PixCorr is the correlation of RGB pixels, SSIM the structural similarity of grayscale images, and PSNR the peak signal-to-noise ratio of RGB pixels.
LPIPS is a distance over AlexNet's five convolutional layers, calibrated to human judgments.
Two-way identification counts how often a reconstruction's features correlate more with its own photograph than with another test photograph of the same dataset, so it is easier on datasets whose photographs differ more from each other.
EfficientNet-B1 and SwAV give correlation distances between embeddings.
Eight of the metrics (all but $r$, PSNR, LPIPS and DINOv2) are those of Brain-Diffuser~\cite{ozcelik2023natural} and MindEye2~\cite{scotti2024mindeye2}, computed with MindEye2's definitions; values can differ from those printed in each paper, which use other image sizes or code.
For our monkeys and the retina, whose reconstructions cover only part of the image, both images are mid-gray outside the covered region, PixCorr, SSIM and PSNR use only that region, and two-way identification excludes other versions of the same photograph.
THINGS-MEG shows each test image 12 times for 500\,ms at 10$^\circ$; ATM reads 271 sensors, recorded at 1,200\,Hz, band-passed at 0.1--40\,Hz and stored at 200\,Hz.
THINGS-EEG2 shows each test image 80 times for 100\,ms at 7$^\circ$, recorded at 1,000\,Hz through a 0.1--100\,Hz filter; ATM and ENIGMA read 63 sensors at 250\,Hz, and Perceptogram 17 at 100\,Hz.
NSD shows each image for 3\,s at 8.4$^\circ$, up to three times (2,770 trials for the 982 test images), and measures the brain every 1.6\,s; its decoders read one response per trial: Brain-Diffuser and MindEye2 from the NSDGeneral voxels (13,931 on average over the four subjects), Takagi from the early and ventral visual regions (12,126), and Brain-IT from about 40,000 voxels per subject.
TVSD shows each test image 30 times for 200\,ms as a 500-pixel square (19$^\circ$ at 25.9 and 26.9 pixels per degree) to 960 working sites in monkey N and 1,024 in F; its multi-unit activity, sampled at 30\,kHz, is low-pass filtered at 200\,Hz and stored at 1\,kHz.
FuzzyAlign and Ciferri et al.\ read it 0--200\,ms after onset, Perceptogram its mean over a 100-ms window per area, and MonkeySee over 133\,ms, at the 656 (N) and 477 (F) sites its code keeps.
Brain2GAN averages 20 presentations into one value per site, over a 100-ms window per area, at monkey N's 960 sites; its 500-pixel images are also 19$^\circ$.
Wu et al.\ decode each image from one presentation, reading 592 to 775 ganglion cells in 1-ms spike bins; the scored images cover 2.3 by 1.5\,mm of retina (11 by 7$^\circ$ at 210\,$\mu$m per degree).
Our monkeys saw each test photograph 60 to 90 times for about 10\,s on 1280-by-720-pixel displays at 37.5 (Monkeys A and L) and 27.7 (Monkeys O and C) pixels per degree; our decoders read the spikes of the units in Table~\ref{tab:sessions} in 120-Hz bins.
Our model trains on all five monkeys, each fold on the photographs outside its test set (19 to 26 in V1, 51 to 62 in LGN); our probes sample at 30\,kHz, Wu et al.'s array at 20\,kHz.
LPIPS compares each reconstruction with its photograph.
For our monkeys and the retina, the reconstruction covers only part of the image, so LPIPS is computed on the largest square inside the covered region; these squares range from 31\,px (retina, flashed) to 612\,px (Monkey O).
LPIPS rewards realistic texture even when the content is wrong, so we read it together with correlation.

Figure~5A lists our four decoders on Monkeys A and O.
All four decoders use the same five folds, split by photograph, so no photograph appears in both training and test.
Parameters count everything used at inference, including frozen pretrained parts; the diffusion count includes the transformer that conditions it.

Figure~5B orders the datasets by the correlation of the methods it shows, lowest first; each method is scored on its own dataset's images, so the order does not rank methods.
Figure~5B shows three held-out photographs for each dataset, at the 90th, 50th and 10th percentiles of correlation averaged over the methods we scored on that dataset, with reconstructions by one or two of them; for this ranking, images larger than 256\,px are first downsampled to 256\,px.
The average is over Brain-IT and MindEye2 for NSD, ATM, ENIGMA and Perceptogram for EEG, ATM alone for MEG, all four TVSD methods, and our transformer and diffusion for Monkeys A and O, on the largest square inside the covered region.
For the retina, it shows the flashed condition, whose reconstructions score highest.
The examples are NSD subject~1 (40\,h), EEG subject~8, MEG subject~1 and TVSD monkey~N.
Figure~\ref{figS:landscape}B shows Brain-Diffuser and Takagi on the same NSD photographs, ranked over Brain-IT and MindEye2 on all 1,000 images.
The TVSD and Brain2GAN examples are development images, so no sealed test image is shown, while their points in Figure~5C are test scores.
The THINGS-MEG test set contains all 100 TVSD test and development images, so the MEG examples come from its other 100 images, while its point in Figure~5C averages all 200.
The retina's examples are cropped to the largest square inside the stimulus.
Fig.~\ref{figS:bands} gives correlations by spatial frequency for every method, and Fig.~\ref{figS:landscape} enlarges Figure~5C with every method named.

%% file: text/appendix_prior.tex
\section{Reconstruction with a \texorpdfstring{$1/f$}{1/f} \TitlePriorTail}
\label{app:derivation}
Equation~\ref{eq:map} reconstructs a scene by choosing the image that fits the decoded patches and is likely under a $1/f$ prior.
This appendix shows where Equation~\ref{eq:map} comes from.
We begin with an easier problem, in which a few pixels of a photograph are observed exactly.
We then replace the observed pixels with patches decoded from neural activity.

\subsection{\TitleFill}
\label{app:einstein}
\begin{figure}[htbp]
    \centering
    \includegraphics[width=340pt]{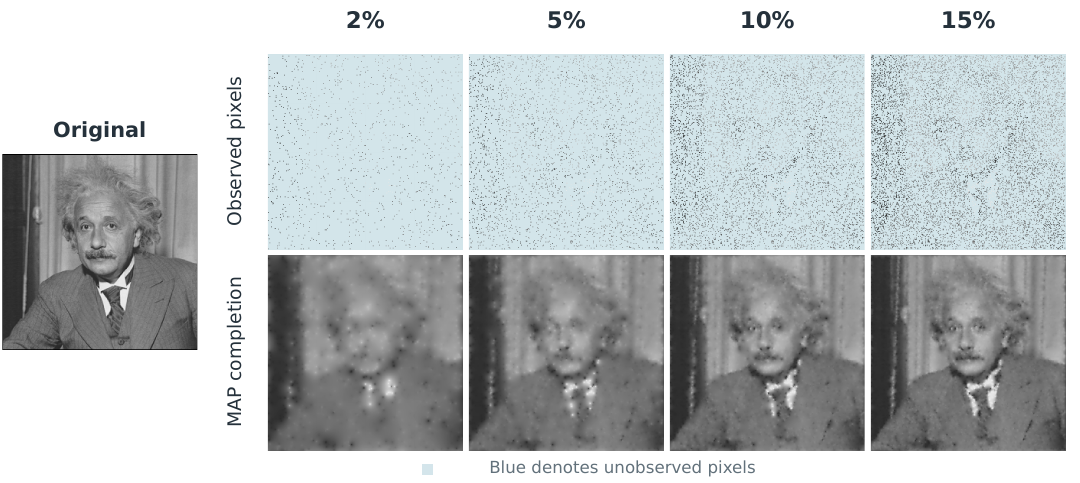}
    \caption{\textbf{Filling in a photograph from a few of its pixels.}
    Left: the original photograph ($256\times256$ pixels).
    Top row: the observed pixels, 2\%, 5\%, 10\% or 15\% of the photograph; unobserved pixels are blue.
    Bottom row: the most probable image given those pixels under the $1/f$ prior (Equation~\ref{eq:einstein-map}), the maximum a posteriori (MAP) completion.
    Each column keeps the observed pixels of the column to its left and adds more.
    All panels share one grayscale range.}
    \label{fig:einstein-inpainting}
\end{figure}

Suppose that we observe only a small fraction of the pixels of a photograph (Fig.~\ref{fig:einstein-inpainting}, top row).
What is the most probable photograph?
Bayes' rule (Equation~\ref{eq:bayes}) gives every candidate image $I$ a probability:
\begin{equation}
    p(I\given\text{observed pixels})\;\propto\;p(\text{observed pixels}\given I)\cdot p(I).
    \label{eq:einstein-bayes}
\end{equation}
The first term on the right is the likelihood.
The observed pixels are exact here, so the likelihood only rules out every image that differs from them.
The second term is the prior, the probability of an image before any pixel is observed.
Among the images that keep the observed pixels, the most probable image is therefore the one that the prior favors most.

\textbf{The $1/f$ prior.}
In natural images, the amplitude at spatial frequency $f$ falls roughly as $1/f$, so the power falls as $1/f^2$~\citep{field1987,olshausenfield2000vision}.
Most of the power is at low spatial frequencies.
We turn this regularity into a prior on the Fourier transform of the image, $\tilde I$.
Each Fourier coefficient $\tilde I(\mathbf f)$ is treated as an independent Gaussian variable with mean zero and expected power $1/(\gamma\lvert\mathbf f\rvert^{2})$.
The probabilities of independent coefficients multiply, so their negative logarithms add:
\begin{equation}
    -\log p(I)=\frac{\gamma}{2}\sum_{\mathbf f}\lvert\mathbf f\rvert^{2}\,\lvert\tilde I(\mathbf f)\rvert^{2}+\text{constant}.
    \label{eq:einstein-prior}
\end{equation}
Here and below, the constant collects the terms that do not depend on the candidate image.
An image is probable under this prior when the sum is small.
A coefficient at a high spatial frequency adds more to the sum than an equally large coefficient at a low spatial frequency, so the prior favors smooth images.
The weight $\gamma$ sets the strength of the prior.
The mean brightness of the image, at $\mathbf f=\mathbf 0$, adds nothing to the sum, so the prior leaves the brightness free.

\textbf{The most probable image} $\hat I$ minimizes the sum in Equation~\ref{eq:einstein-prior} while keeping every observed pixel:
\begin{equation}
    \hat I=\arg\min_I\sum_{\mathbf f}\lvert\mathbf f\rvert^{2}\,\lvert\tilde I(\mathbf f)\rvert^{2}\quad\text{with the observed pixels of $I$ held fixed}.
    \label{eq:einstein-map}
\end{equation}
The weight $\gamma$ drops out, because it scales the sum of every candidate image by the same factor.
We find $\hat I$ by gradient descent on the unobserved pixels.
Figure~\ref{fig:einstein-inpainting} shows $\hat I$ for one photograph.
From the smallest fraction of pixels, the prior recovers the outline of the head and shoulders.
From the largest fraction, the face is easy to recognize.
The fine detail of the photograph does not return, because a $1/f$ spectrum is all that this prior says about natural images.
This demonstration uses no neural activity and no learned model.

\subsection{\TitlePatches}
\label{app:ridge-mosaicing}
Reconstructing a scene from neural activity is the same problem with different observations.
The observations are now the calibrated patches $\mathbf p_1,\dots,\mathbf p_K$, which ridge regression decodes from spikes and gaze places in the scene (Section~\ref{sec:ridge}).
Decoded patches differ from observed pixels in two ways: a decoded patch is noisy, and many patches land on the same pixel.

\textbf{Likelihood.}
We model each patch pixel $p_k(\mathbf x)$ as the pixel $I(\mathbf x)$ of the scene plus independent Gaussian noise.
The variance of the noise is $1/\pi_k$, because the reliability $\pi_k$ of a decoder is the inverse of its mean squared error.
The likelihood of all the patches is then a product of Gaussians, one for every patch pixel, and its negative logarithm is a sum:
\begin{equation}
    \begin{aligned}
    -\log p(\text{patches}\given I)
    &=\frac12\sum_{\mathbf x}\sum_{k\ni\mathbf x}\pi_k\bigl(p_k(\mathbf x)-I(\mathbf x)\bigr)^{2}+\text{constant}\\
    &=\frac12\sum_{\mathbf x}w(\mathbf x)\bigl(I(\mathbf x)-M(\mathbf x)\bigr)^{2}+\text{constant}.
    \end{aligned}
    \label{eq:mosaic-likelihood}
\end{equation}
The second line holds because, at each pixel, the weighted squared errors against all the patches equal one squared error against their weighted average, plus a term that does not depend on the image.
The weighted average is the patch mosaic $M(\mathbf x)$ of Equation~\ref{eq:mosaic}, and its weight is the overall reliability $w(\mathbf x)$.
The patches therefore enter the reconstruction only through $M$ and $w$.

\textbf{Posterior.}
Bayes' rule multiplies this likelihood by the $1/f$ prior of Equation~\ref{eq:einstein-prior}, so the two negative logarithms add:
\begin{equation}
    -\log p(I\given\text{patches})=\frac12\sum_{\mathbf x}w(\mathbf x)\bigl(I(\mathbf x)-M(\mathbf x)\bigr)^{2}+\frac{\gamma}{2}\sum_{\mathbf f}\lvert\mathbf f\rvert^{2}\,\lvert\tilde I(\mathbf f)\rvert^{2}+\text{constant}.
    \label{eq:mosaic-posterior}
\end{equation}
The most probable image $\hat I$ minimizes the right-hand side.
This minimization is Equation~\ref{eq:map}, since the shared factor of one half does not move the minimum.
Unlike the observed pixels of Appendix~\ref{app:einstein}, a decoded pixel is allowed to change.
Where many reliable patches land, $w(\mathbf x)$ is large and $\hat I$ stays close to the mosaic.
Where few patches land, the prior smooths the noise.
Pixels that no patch reaches have $w(\mathbf x)=0$ and are left to the prior.

\textbf{Patches and calibration.}
The patches are $21\times21$ grayscale pixels per V1 window, and $13\times13$ samples spanning $0.75^\circ$ in each of three color channels per LGN receptive-field group.
Windows are back-to-back 100-ms stimulus windows, with spikes counted 16.7\,ms later and placed at the median gaze.
For each decoder, the ridge penalty $\lambda$ is chosen by cross-validation over training photographs.
The decoder is fitted on the training photographs not reserved for calibration.
The two gains and the reliability of the decoder (Section~\ref{sec:ridge}) are then measured on the reserved photographs, separately for each color channel in LGN.

\textbf{Mosaic.}
Each calibrated patch is spread onto the pixel grid of the scene by bilinear interpolation.
Every patch carries the same total weight before the weight is multiplied by the reliability of its decoder.
In LGN, the weight of a group is also multiplied by the mean inverse number of retained groups that its neurons belong to, so that overlapping groups do not count the same neurons twice.
The patch mosaic $M(\mathbf x)$ is defined wherever $w(\mathbf x)>0$.

\textbf{Color.}
In LGN, both sums of Equation~\ref{eq:map} also run over the three color channels.
The prior term of each channel is multiplied by the ratio of luminance power to the power of that channel in the training photographs.
One weight $\gamma$ serves all three channels.

\textbf{Prior weight and solver.}
The squared frequency $\lvert\mathbf f\rvert^{2}$ is measured relative to its average over the whole Fourier grid, which fixes the scale of $\gamma$.
The weight $\gamma$ is chosen per animal and fold from $\{2,4,8,16,32\}$ by the mean correlation on validation photographs inside the training set.
Equation~\ref{eq:map} is quadratic in the image, so conjugate gradients solve it, with fast Fourier transforms for the prior term, to a relative residual of $2\times10^{-6}$.
The Fourier transforms treat the scene as periodic.
The result is clipped to $[0,1]$ for display and scoring.

\textbf{Held-out photographs.}
Decoder fitting, calibration and these per-fold choices exclude each fold's test photographs.

%% file: text/appendix_recordings.tex
\section{Animals and Recordings}
\label{app:recordings}
We decode recordings from five monkeys: three marmosets recorded in V1 (Monkeys A, L and L2) and two macaques recorded in the LGN (Monkeys O and C).
Section~\ref{app:rec-al} reproduces the recording methods of \citet{ressmeyer2026active}{} for the two marmosets of that study, Monkeys A and L.{}
Sections~\ref{app:rec-l2} and~\ref{app:rec-lgn} describe the other three monkeys in the same format, and Section~\ref{app:rec-units} lists the sessions and units used for decoding.

\subsection{Marmosets A and L (V1)}
\label{app:rec-al}
\paragraph{Animals and surgical preparation.}
Two adult male common marmosets (\textit{Callithrix jacchus}), monkeys L and A, contributed the recordings analyzed here.
The monkey L recordings were collected as part of the dataset described by Yates et al.~\cite{yates2023}.
The monkey A recordings were collected subsequently on the same preparation and acquisition system, with the modifications noted below.
The fixated flashed-image condition was not part of that report.
All surgical and experimental procedures were approved by the University Committee on Animal Resources at the University of Rochester{} and conformed to US National Institutes of Health guidelines.
Each animal was implanted with a titanium head post and a recording chamber positioned over the foveal representation on the lateral surface of V1~\cite{yates2023}.
The electrode arrays were lowered into V1 through a craniotomy inside the chamber with a removable microdrive~\cite{yates2023}.
We use 16 sessions from Monkey L, recorded between December 2019 and March 2020, and 14 from Monkey A, recorded between February and August 2022.

\paragraph{Eye tracking and stimulus presentation.}
Animals were head-fixed during recording and viewed the display through a hot mirror.
Eye position was measured with a digital dual Purkinje image eye tracker~\cite{yates2023,wu2023dpi,ressmeyer2026}.
Visual stimuli were generated in MATLAB using Psychtoolbox~\cite{brainard1997psychophysics} and presented with a ProPixx projector (VPixx Technologies).
The projector showed 1280$\times$720 pixels at a refresh rate of 240\,Hz, and the viewing distance of 79\,cm gave 37.5 pixels per degree.
Stimulus and electrophysiology clocks were synchronized with a Datapixx interface, which transmitted event words to the Open Ephys acquisition system.
Stimulus code is available at \url{https://github.com/jcbyts/MarmoV5}.
\paragraph{Fixated flashed-image condition.}
Animals maintained fixation while viewing a repeated sequence of whitened natural images.
Images subtended 3$^\circ$ of visual angle and were presented at 20\,Hz.
Analyses were restricted to periods in which valid eye-position estimates remained within 0.5$^\circ$ of the fixation point, a 1$^\circ$-diameter fixation window.
Repeated presentations of the same sequence provided matched stimulus epochs with naturally varying fixational eye trajectories.

\paragraph{Free-viewing stimulus conditions.}
Each session also included three free-viewing conditions taken from the foraging paradigm described previously~\cite{yates2023}{}.
No fixation was required in these conditions.
The animal freely viewed the display while stimuli were presented continuously.
With Gabor noise and gratings, small Gabor or face targets appeared intermittently at varying screen locations to encourage saccades.
Eye position was tracked throughout, and the stimulus falling on each receptive field was reconstructed offline in retinal coordinates.
In that study, these conditions were not used for the variability decomposition.
They provided the receptive-field estimates used to select units for analysis, the training data for the encoding model, and the natural free-viewing fixations analyzed in Figure~4 of \citet{ressmeyer2026active}.

\textit{Gabor noise.} A dense field of Gabor patches covered the display, and a new set of patches was drawn on each noise frame, with the orientation, phase, spatial frequency, size, and position of every patch drawn independently at random.
This condition drove units across a wide range of orientations and spatial frequencies, and provided the data used to compute spike-triggered averages and spike-triggered energies.

\textit{Gratings.} A full-field sinusoidal grating was presented at fixed phase and contrast, with its orientation and spatial frequency redrawn on each noise frame.

\textit{Natural images.} A single natural image spanning the display was presented on each trial and freely viewed.
Each image was shown in grayscale for 10\,s, with 2\,s between images, and no targets appeared in this condition~\cite{yates2023}.
Both monkeys saw the same 30 images, and the decoders use 27 of them.
In that study, fixations extracted from this condition provided the eye-position distributions and local image measurements analyzed in its Figure~4.

\paragraph{Electrophysiology and spike sorting.}
Neural activity was recorded with multisite silicon electrode arrays (NeuroNexus).
Each array contained two shanks separated laterally by 200\,$\mu$m, with 32 recording sites per shank at 35\,$\mu$m spacing.
Arrays used for monkey A were coated with poly(3,4-ethylenedioxythiophene) (PEDOT) to reduce impedance and improve signal quality~\cite{cui2003pedot}.
Wideband signals were amplified and digitized at 30\,kHz with Intan headstages using Open Ephys.
Recordings from monkey L were processed and spike sorted with Kilosort 2 following Yates et al.~\cite{yates2023}, including manual curation in Phy.
Recordings from monkey A were processed with an updated pipeline and spike sorted with Kilosort 4~\cite{roth2026quality}.
Both single-unit and multi-unit clusters were considered for analysis.
The ridge decoders read every cluster: 2,315 for Monkey A and 967 for Monkey L.
The transformer reads the clusters with enough spikes during natural-image viewing, at least 500 for Monkey A (2,230 clusters) and at least 100 for Monkey L (897); Table~\ref{tab:sessions} counts these.

\subsection{Marmoset L2 (V1 and V2)}
\label{app:rec-l2}
\paragraph{Animals and surgical preparation.}
Monkey L2, a male common marmoset (\textit{Callithrix jacchus}), contributed 14 recording sessions collected at UCLA between July 2025 and March 2026.
Surgical and experimental procedures were approved by UCLA's Chancellor's Animal Research Committee under protocol ARC-2022-085.
Monkey L2 had a titanium head post and burr holes over V1 and V2 in the right hemisphere.

\paragraph{Eye tracking and stimulus presentation.}
Eye position was measured with OpenIrisDPI~\cite{ressmeyer2026}.
Stimuli were generated with the same MATLAB code and followed the Gabor-noise, fixated flashed-image, and freely viewed natural-image paradigms described above.
Stimuli were presented on a 2560$\times$1440 ASUS PG27AQN LCD at 240\,Hz.
Viewing distance was 67.5\,cm in July--August 2025 (53.7 pixels per degree), 62\,cm in November 2025 (49.9 pixels per degree), and 63.5\,cm in December 2025 and March 2026 (50.9 pixels per degree).
Monkey L2 viewed 103 grayscale photographs selected from set 9 of the University of Texas at Austin Natural Scene Collection~\cite{geisler2011statistics}.
Each photograph was shown for approximately 10\,s, with 2\,s between photographs.

\paragraph{Electrophysiology and spike sorting.}
Neural activity was recorded with one or two standard Neuropixels~1.0 probes~\cite{jun2017neuropixels} per session.
Probes were inserted acutely through the right-hemisphere burr holes using a procedure adapted from \citet{liska2023running}.
The dura was nicked and the opening filled with saline.
Each probe was lowered at 200\,$\mu$m/min, with 100\,$\mu$m of retraction after each 500\,$\mu$m of insertion to avoid sustained tissue compression.
Whole-probe LFP was monitored during insertion.
Probes were allowed to settle for a further 30\,min after placement.
Signals were acquired at 30,000 samples/s using Neuropixels headstages and IMEC PXIe hardware with SpikeGLX, and spikes were sorted with Kilosort~4~\cite{pachitariu2024kilosort4}.
The 12,184 units in Table~\ref{tab:sessions} were pooled across Monkey L2's 14 sessions for model training without restricting them by a V1 or V2 label.

\subsection{Macaques O and C (LGN)}
\label{app:rec-lgn}
\paragraph{Animals and surgical preparation.}
Two adult rhesus macaques (\textit{Macaca mulatta}), one male (Monkey O) and one female (Monkey C), contributed recordings from the LGN.
They are monkeys O and C of \citet{ressmeyer2026}, whose methods we summarize here.{}
All surgical and experimental procedures were approved by the Institutional Animal Care and Use Committee at the University of Washington{} and conformed to US National Institutes of Health guidelines.
Each monkey was surgically implanted with a titanium head post and a recording chamber (Crist Instruments) stereotaxically positioned over the LGN.
We use 27 sessions from Monkey O, recorded between January 2024 and February 2026, and 23 from Monkey C, recorded between September and December 2024.

\paragraph{Eye tracking and stimulus presentation.}
The monkeys sat in a primate chair (Crist Instruments) in a dark room and viewed a rear-projected display (VPixx PROPixx)~\cite{ressmeyer2026}.
They sat about 61\,cm from the screen, and the projected image was about 51\,cm wide.
In all 50 sessions, the display had 1280$\times$720 pixels at 27.7 pixels per degree and a 240-Hz refresh rate.
Both eyes were imaged at 500\,Hz with OpenIrisDPI, an open-source digital dual Purkinje image eye tracker~\cite{ressmeyer2026}.
Gaze was calibrated in each session with a quadratic fit to fixations on targets at known positions, and eye-tracker samples were synchronized to the neural recording clock.
The decoders use one eye per session: the right eye in 44 sessions and the left eye in 6.
For each session we took the eye that gave the sharper receptive-field maps, measured by the signal-to-noise ratio of the spike-triggered averages of the units mapped with both eyes.

\paragraph{Natural-image foraging.}
The monkeys viewed 70 color photographs, each shown for about 10\,s at a time.
The photographs are calibrated images from the UPenn Natural Image Database~\cite{tkacik2011natural}.
They were displayed so that the cone activations were proportional to those of the calibrated images, using the spectra of the projector measured with a PR-650 spectroradiometer (Photo Research).
In every session, the photographs were shown during a foraging task: small Gabor targets (0.3$^\circ$ in radius, 5 cycles per degree) appeared on the photograph, and the monkey fixated each one to receive a reward.
The contrast of the targets was 80\% in 39 sessions and between 20\% and 100\% in the other 11.
In thirty-three sessions, the natural-image trials ran in blocks of their own; in the other seventeen, they were interleaved with white-noise and blank trials.
A target stayed on until the monkey had fixated it for a required time, which we estimated from the eye traces: about 0.6\,s in the block-wise sessions (0.1\,s in the first two) and 0.4 to 1.0\,s in the interleaved ones.
Targets followed one another about every 1.3\,s in the block-wise sessions and every 1.4\,s in the interleaved ones.
The block-wise sessions, all recorded in 2024, used an earlier color rendering in which the S-cone signal was about five times too low, so their photographs look warmer.
The interleaved sessions, recorded from Monkey O between December 2025 and February 2026, used the corrected rendering.

\paragraph{Electrophysiology and spike sorting.}
Neural activity was recorded from the LGN with high-density silicon probes (Neuropixels NHP 1.0 long) and acquired at 30\,kHz with Open Ephys~\cite{ressmeyer2026}.
Each probe was advanced through a guide tube with a hydraulic microdrive until it reached the LGN, which was confirmed with a hand-held flickering light.
Spikes were sorted with Kilosort~2.5~\cite{pachitariu2024kilosort4}, which yielded 553 units per session on average (range 281--826).
The decoders use a subset of these units: 199 per session on average for Monkey O and 121 for Monkey C.
These are the units whose receptive field could be mapped with white noise: the spike-triggered average or the spike-triggered energy, computed with the corrected gaze, had a signal-to-noise ratio of at least 3.

\subsection{Sessions and Units Used for Decoding}
\label{app:rec-units}
Table~\ref{tab:sessions} lists the sessions and units the decoders use.
In total, they cover 94 sessions and 23,452 units.

{\centering\small
\captionof{table}{\textbf{Sessions and units used for decoding.} Units per session are means over sessions.}
\label{tab:sessions}
\begin{tabular}{@{}lllrrr@{}}
\toprule
Monkey & Species & Area & Sessions & Units & Units per session \\
\midrule
A & Marmoset & V1 & 14 & 2,230 & 159 \\
L & Marmoset & V1 & 16 & 897 & 56 \\
L2 & Marmoset & V1, V2 & 14 & 12,184 & 870 \\
O & Macaque & LGN & 27 & 5,360 & 199 \\
C & Macaque & LGN & 23 & 2,781 & 121 \\
\midrule
Total & & & 94 & 23,452 & 249 \\
\bottomrule
\end{tabular}\par}

%% file: text/appendix_offset.tex
\section{The Windowed Offset Term}
\label{app:offset}
In the query readout (Section~\ref{sec:transformer}), each query pools the content of its nearest tokens with attention weights.
Pooling alone does not tell the query where a token's content lies relative to the query's own block of pixels.
Each of the three local readout layers therefore has a second term, which shifts every token's content by the offset between that token and the query before pooling it.
The token's content is mapped linearly to 256 complex numbers, each paired with a fixed spatial frequency.
Multiplying each number by the phase that its frequency gives the offset shifts the content by that offset, just as translating an image changes the phases of its Fourier transform.
The frequencies are drawn once, uniformly from all those with periods longer than two pixels of the decoded scene.
The shifted content is then weighted by a Gaussian of the offset's length, with a standard deviation of 6 pixels of the $180\times320$ decoded scene (24 pixels of the photograph at full size), so a token places content only near where its receptive fields landed.
The weighted contents are pooled with the layer's attention weights and added to its output through a linear map that starts at zero, so training begins from the model without the term.
The term adds 0.8M parameters and is used in training and at test time.